\documentclass[conference]{IEEEtran}
\usepackage{times}

\usepackage[numbers]{natbib}
\usepackage{multicol}
\usepackage[bookmarks=true]{hyperref}
\usepackage{multirow}
\usepackage{array}
\usepackage{hyperref}
\usepackage{url}
\usepackage{float}
\usepackage{tabularx}  
\usepackage{booktabs}
\usepackage{xcolor} 
\usepackage{colortbl}
\usepackage{mathtools}
\usepackage{amsfonts}  
\usepackage{graphics}
\usepackage[pdftex]{graphicx}
\usepackage{wrapfig}
\usepackage{caption}
\usepackage{color}
\usepackage{dsfont}
\usepackage[]{mdframed}
\usepackage{float}
\usepackage[ruled,vlined,linesnumbered]{algorithm2e}
\usepackage{xparse}
\usepackage{amsmath}
\usepackage{bm}
\usepackage{mathtools}
\usepackage{amssymb}
\usepackage{amsthm}
\usepackage{amssymb}

\usepackage{booktabs}
\usepackage{epigraph}

\usepackage{xcolor} 
\usepackage{hyperref} 
\hypersetup{
    colorlinks=true,
}

\usepackage{xargs} 
\usepackage[colorinlistoftodos,prependcaption,textsize=tiny]{todonotes}
\newcommandx{\wrn}[2][1=]{\todo[linecolor=red,backgroundcolor=red!25,bordercolor=red,#1]{#2}}
\newcommandx{\cmt}[2][1=]{\todo[linecolor=blue,backgroundcolor=blue!25,bordercolor=blue,#1]{#2}}

\IEEEoverridecommandlockouts

\begin{document}
\makeatletter
\let\@oldmaketitle\@maketitle
\renewcommand{\@maketitle}{\@oldmaketitle
  \begin{center}
  \captionsetup{type=figure}
  \setcounter{figure}{0}
  \includegraphics[width=1.0\textwidth,height=0.28\textheight]{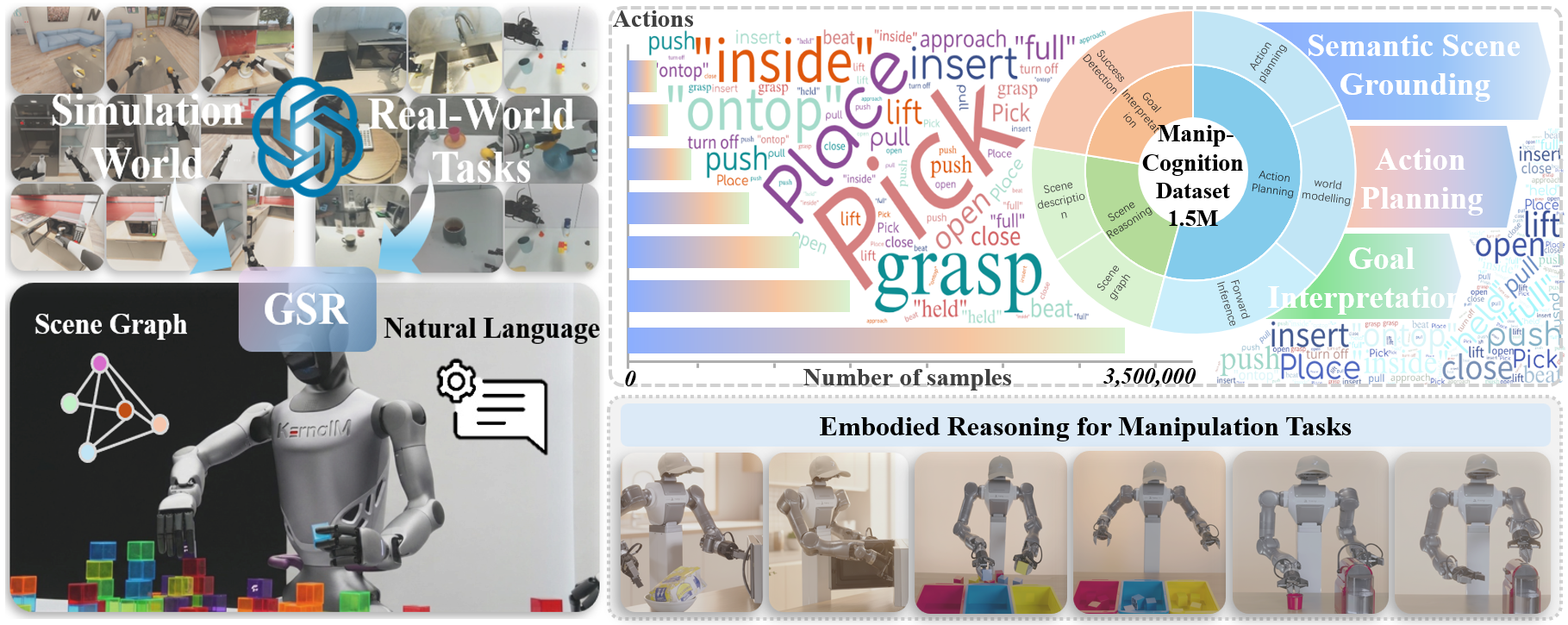}
    \captionof{figure}{In this study, we introduce GSR and the Manip-Cognition Dataset. GSR advances robotic manipulation through rational reasoning ability, while Manip-Cognition provides 1.6M samples spanning scene grounding, goal interpretation, and action planning across a wide range of atomic manipulation skills. } 
    \label{fig:overall of data}
    \end{center}
}

\title{GSR: Learning Structured Reasoning for \\ Embodied Manipulation}

\def\cameraready{0}  

\ifx\cameraready\undefined
    \author{
    Anonymous Submission
    }
\else
    \author{
    Kewei Hu$^{\#}$,
    Michael Zhang$^{\#}$,
    Wei Ying,
    Tianhao Liu,
    Guoqiang Hao,
    Zimeng Li,\\
    Wanchan Yu,
    Jiajian Jing,
    Fangwen Chen,
    Hanwen Kang$^{*}$\\
    \url{https://klmmotion.github.io/gsr.github.io/}
    }
\fi

\maketitle

\begin{abstract}
Despite rapid progress, embodied agents still struggle with long-horizon manipulation that requires maintaining spatial consistency, causal dependencies, and goal constraints. A key limitation of existing approaches is that task reasoning is implicitly embedded in high-dimensional latent representations, making it challenging to separate task structure from perceptual variability.
We introduce Grounded Scene-graph Reasoning (GSR), a structured reasoning paradigm that explicitly models world-state evolution as transitions over semantically grounded scene graphs. By reasoning step-wise over object states and spatial relations, rather than directly mapping perception to actions, GSR enables explicit reasoning about action preconditions, consequences, and goal satisfaction in a physically grounded space.
To support learning such reasoning, we construct Manip-Cognition-1.6M, a large-scale dataset that jointly supervises world understanding, action planning, and goal interpretation. Extensive evaluations across RLBench, LIBERO, GSR-benchmark, and real-world robotic tasks show that GSR significantly improves zero-shot generalization and long-horizon task completion over prompting-based baselines. These results highlight explicit world-state representations as a key inductive bias for scalable embodied reasoning.
\end{abstract}

\IEEEpeerreviewmaketitle

\section{Introduction}
\IEEEPARstart{E}{mbodied} learning has become an important approach for building robotic systems that integrate perception, reasoning, planning, and execution. 
Despite notable progress, existing methods continue to struggle with adaptation to real-world variability and with completing long-horizon tasks under diverse semantics \cite{black2024pi_0}. 
These challenges are closely related to how models interpret observations, understand language instructions, and connect high-level intentions to low-level actions. 
Many current approaches do not explicitly separate high-level task reasoning. 
As a result, task logic often becomes entangled with visual variability, making models overly sensitive to environmental changes. 
This limitation becomes evident when agents are required to operate beyond training data to solve tasks that demand planning over long sequences \cite{zhao2025cot}.

An embodied agent relies on two core capabilities: embodied reasoning and action control. 
The embodied reasoning infers intention-oriented sequenced action (e.g., “\textit{lean the refrigerator}” or “\textit{place the cup on the plate}”) from perceptual inputs \cite{chen2024roboscript}. 
This requires an agent to perceive the environment, understand the intent expressed in human instructions, and generate correctly sequenced actions to complete the task. How the world is represented plays a critical role in this process. World representations are influenced by multiple factors, including object appearance, spatial relationships, and task semantics. Even within the same scene, different goals can lead to entirely different action sequences; conversely, under the same goal, changes in the scene configuration may also require different behaviors.
However, current implicit neural network models, such as convolutional neural networks (CNNs) and Vision Transformers (ViTs), struggle to extract such high-level semantic and relational information needed for effective reasoning. As a result, their ability to generalize across scenes and tasks remains limited \cite{radford2021learning}. Consequently, transferring to new tasks or adapting to even minor environmental changes often requires substantial additional data collection and retraining.
Classical Task and Motion Planning (TAMP) frameworks offer structured solutions for such planning ability but typically rely on predefined symbolic representations (e.g., PDDL and BDDL). These approaches are often difficult to scale to complex, open-ended environments with high variability \cite{ji2025robobrain}. 
Motivated by these limitations, this study explores: \textbf{Can embodied agents leverage structured intermediate representations to enhance reasoning abilities general manipulation tasks?}

To this end, we introduce \textbf{Grounded Scene-graph Reasoning (GSR)}, an embodied reasoning framework based on the principle that agents should plan over abstract world representations rather than raw visual information. GSR leverages semantically grounded scene graphs to extract stable causal structures from observations and explicitly separates high-level conceptual reasoning from low-level action execution. This design enables persistent and task-transferable capabilities, allowing flexible composition of sequenced action and robust adaptation across tasks. 
To train GSR, we construct the \textbf{Manip-Cognition-1.6M}, which provides joint supervision over world understanding, intention interpretation, and action planning across a diverse set of manipulation tasks.

The remainder of this paper is structured as follows: 
Section \ref{sec: method} introduces the methods; 
Section \ref{sec: exp} presents the experimental setup and results; 
Section \ref{sec: conclusion} concludes the paper.
 
\section{Related Works}
\subsection{World Representations for Robot Learning}

Current emboided-learning models generally adopt either end-to-end or hierarchical modular designs. 
End-to-end approaches, including RT-2 \cite{zitkovich2023rt}, OpenVLA \cite{kim2024openvla}, and $\pi_0$ \cite{black2024pi_0}, leverage large-scale data to directly map raw visual observations to actions, exhibiting strong scaling behavior. 
However, operating on high-dimensional sensory inputs introduces a representation bottleneck \cite{ai2025review}, in which task-irrelevant visual variability becomes entangled with underlying physical dynamics. The lack of explicit state abstraction biases learning toward appearance-driven correlations rather than causal structure, leading to degraded generalization in out-of-domain settings and long-horizon manipulation tasks \cite{brohan2022rt}.
To mitigate these issues, hierarchical VLA such as RT-H \cite{belkhale2024rt} decompose decision making into high-level planning and low-level control. While this improves modularity, many such approaches still rely on latent embeddings that provide limited spatial and semantic supervision and lack explicit inductive bias for physical structure, resulting in representations that remain sensitive to perceptual variation\cite{yang2023foundation}.
These limitations suggest that architectural decomposition alone is insufficient, and that robust generalization fundamentally depends on the choice of information used to mediate perception, reasoning, and action.

Effective representations introduce inductive bias to separate task structure from perceptual variability \cite{ai2025review}.
Pixel-level and latent representations, while expressive and widely used in end-to-end VLA models \cite{zitkovich2023rt}, lack explicit structural priors for physical interaction and compositional reasoning.
Recently, scene graphs have emerged as a structured and hierarchical representation in robotic applications, explicitly abstracting objects, attributes, and their relations \cite{li2024scene}. 
Beyond compact scene encoding, they provide a language-aligned causal abstraction that links spatial understanding to action reasoning and planning \cite{zhu2024vision}. 
By exposing object states and relational predicates in symbolic form, scene graphs enable models, particularly language-based reasoners, to infer action consequences, preconditions, and dependencies across time. 
This causal grounding supports long-horizon planning by allowing agents to reason over multi-step actions, while bridging low-level perception with high-level linguistic concepts to enable generalizable reasoning (e.g., inferring that a cup is inside a microwave).

\begin{figure*}
    \centering
    \includegraphics[width=1.0\textwidth,height=0.42\textheight]{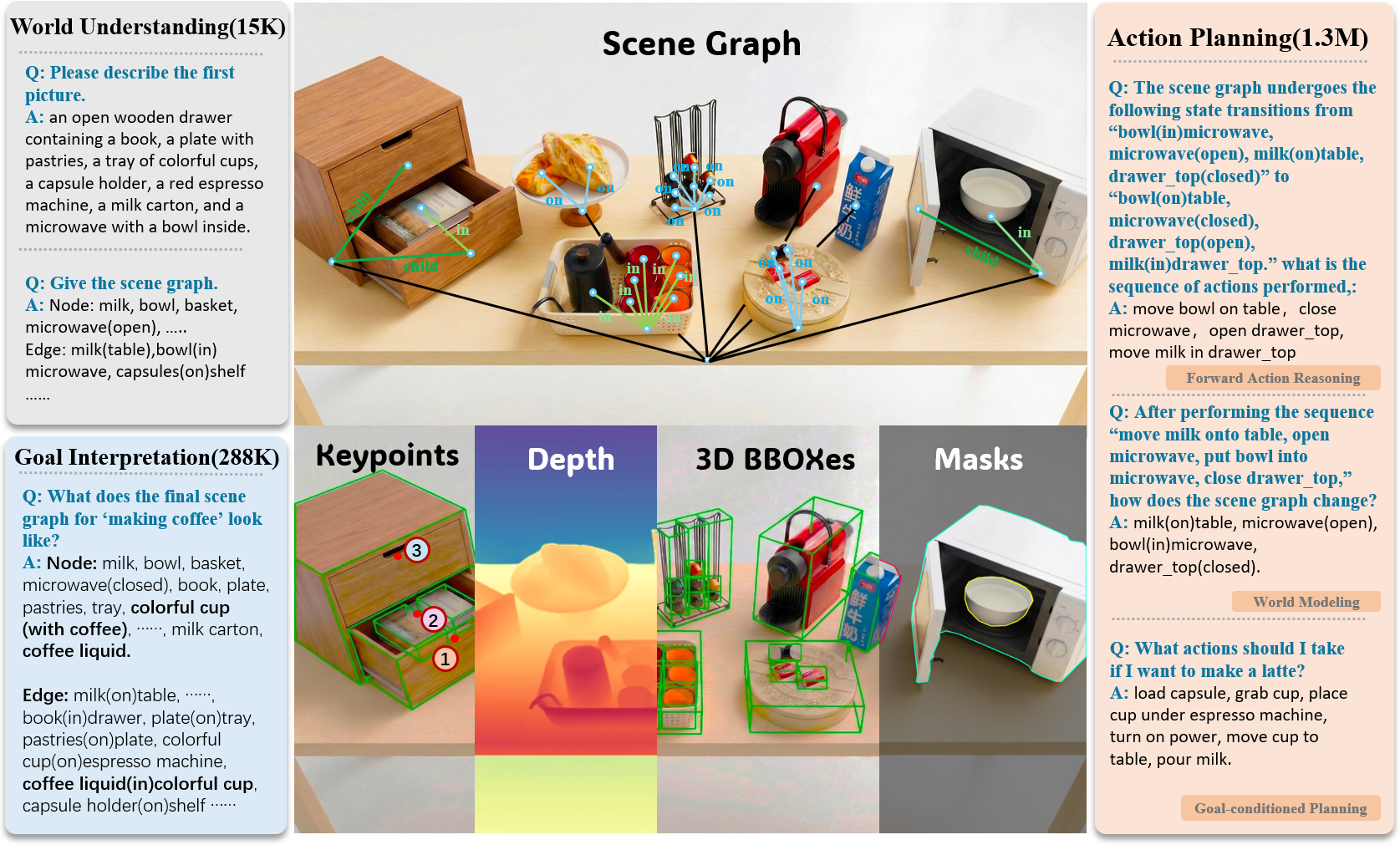}
    \caption{Examples of the Manip-Cognition data including scene grounding data, planning data and goal interpretation data}
    \label{fig:dataset}
\end{figure*}

\subsection{Relation to Recent Works}
Our work is related to prompting-based spatial reasoning approaches such as VoxPoser \cite{huang2023voxposer}, ReKep \cite{huang2024rekep}, ConceptGraphs \cite{gu2024conceptgraphs}, and ENACT \cite{wang2025enact}. VoxPoser and ReKep translate language instructions into motion-level constraints or relational keypoints, but task semantics and world states remain implicitly encoded. ConceptGraphs and ENACT leverage scene graphs for perception, mapping, or auxiliary planning support; however, decision making is still driven by prompt-based or implicit reasoning without an explicit model of world-state transitions.
In contrast, GSR treats the grounded scene graph as the \textbf{primary state space for decision making} and formulates embodied manipulation as explicit reasoning over object-centric state transitions. GSR jointly learns forward state evolution, goal-state inference, and next-action reasoning within a unified embodied framework, enabling step-wise modeling of action preconditions, causal effects, and goal satisfaction. This distinguishes GSR from prior scene-graph-based systems that use graphs as auxiliary representations rather than as the core substrate for embodied reasoning.

\section{Methodology} \label{sec: method}

\subsection{Overview of GSR framework} 

\subsubsection{World Representation}
Given an RGB-D observation $\mathcal{I}$, we construct a 3D scene graph $M_{sg} = (O_t, E_t)$ to encode the workspace and object states, where $O_t = \{o_j\}_{j=1,...,J}$ denotes the set of objects and $E_t = \{e_k\}_{k=1,...,K}$ denotes the set of relational edges.
Figure \ref{fig:dataset} illustrates the transformation process from raw visual input to a structured representation and the resulting scene graph. Each object $o_j$ is represented as a structured entity composed of functional keypoints and, when applicable, articulated child components. For example, a \textbf{mug} includes a "\textit{functional keypoint}" corresponding to its \textbf{handle}, while a \textbf{cabinet} is modeled as an articulated object with "\textit{multiple child elements}" such as \textbf{drawers}. Edges $e_k$ encode spatial relations between object pairs, capturing predicates such as \textit{on}, \textit{inside}, or \textit{adjacent} to (e.g., a \textbf{mug} \textit{on} a \textbf{table}). Additional examples of scene graphs across diverse task scenarios are provided in Appendix~\ref{Appendix: VFM}.

\subsubsection{Model Structure}
GSR is a fine-tuned Large Language Model (LLM) designed to perform commonsense reasoning over scene-graph representations.
To apply GSR in a physical embodiment, we integrate it with a perception front-end and a action expert back-end. 
The physical system consists of two components: a perception–reasoning module for decision making, and an action expert that executes low-level control. 
The perception–reasoning module constructs scene graphs from raw observations and enable GSR reasons over these information to generate sequences of actions. 
To construct scene graphs, an \textbf{Vision Foundation Model (VFM)} (see Appendix~\ref{Appendix: VFM}) is applied.
Action expert leverages a meta-skill library, is detailed in Section \ref{sec: real_robot}.

\subsection{Manip-Cognition Data Engine}
This section introduces the \textbf{Manip-Cognition-1.6M}, which is designed around three core competencies: scene-to-graph understanding, action planning, and goal interpretation, as illustrated in Figure ~\ref{fig:dataset}. 
By processing raw visual demonstrations into structured task trajectories, the data engine enables models to generate rational, sequential actions for completing long-horizon, instruction-driven manipulation tasks. 
\textbf{Besides, we apply systematic data augmentation to the raw trajectories, resulting in a training corpus of 1.6M samples.} 
Detailed data processing procedures are provided in Appendix~\ref{app:data_details}.

\subsubsection{World Understanding Data}
This data component trains the model to abstract visual observations into semantically grounded Scene Graphs ($SG$). Sourced from high-quality interaction benchmarks including \textbf{Behavior-1K \cite{li2024behavior}} and \textbf{Enact \cite{wang2025enact}}, this data teaches the model to interpret structured SGs and learn extracting spatial relations from natural language descriptions. The final representation takes the form of standardized $\text{Text} \rightarrow \text{SG}$ pairs. Following linguistic and structural augmentation, \textbf{this subset comprises 15k grounding pairs}, providing the necessary supervision for the model to bridge the gap between perception and symbolic reasoning.

\subsubsection{Action Planning Data} 
This dataset is designed to train models to capture causal relations between objects and actions, which are essential for action sequencing and reasoning under diverse spatial constraints. 
We decompose this capability into three complementary reasoning tasks: 

(a) \textbf{Forward Action Reasoning}: This task focuses on goal-conditioned planning and causal inference. Unlike simple one-step imitation, the model is trained to infer the underlying action sequences responsible for observed or desired state transitions. Given an initial state $SG_t$ and a target state $SG_{t+n}$ (where $n \ge 1$), the model must predict the multi-step action $\mathcal{A} = \{A_t, A_{t+1}, \dots, A_{t+n-1}\}$ required to bridge the gap.
\begin{equation}
    (SG_t, SG_{t+n}) \rightarrow \{A_i\}_{i=t}^{t+n-1}
\end{equation}

(b) \textbf{World Modeling}: This task functions as an internal physics engine, training the agent to predict the environmental consequences of its actions. By analyzing consecutive scene graph pairs within task trajectories, we extract the structural variations in edges ($\Delta SG_{\text{edge}}$) between the pre-action state ($SG_t$) and the post-action state ($SG_{t+1}$). The model is trained to predict this $\Delta SG_{\text{edge}}$ (the resulting state change) given the current $SG_t$ and a specific candidate Action $A_t$.
\begin{equation}
    (SG_t, A_t) \rightarrow \Delta SG_{\text{edge}}
\end{equation}

(c) \textbf{Goal-conditioned Planning}: This dataset aligns high-level linguistic instructions with the structured world state. To ensure the agent can execute tasks based on user intent, the model is trained to predict the immediate next symbolic action $A_t$ required to progress toward a textual task goal $G$, given the current environmental configuration $SG_t$.
\begin{equation}
    (SG_t, G) \rightarrow A_t
\end{equation}

\noindent \textbf{After augmentation, the Planning Dataset comprises a total of 1.3M samples, covering 6k trajectories}.

\subsubsection{Goal Interpretation Data} 
This dataset is designed to instill a deep understanding of task requirements and end-state conditions, enabling the model to internalize "what a completed task looks like" in a structured space. Unlike step-by-step world modeling, this capability focuses on long-horizon goal dreaming: predicting the final environment configuration ($SG_{goal}$) given a current state ($SG_t$) and a natural language instruction. 

\begin{equation}
    (SG_t, \text{Goal Instruction}) \rightarrow SG_{goal}
\end{equation}
This provides the agent with a stable semantic target to guide multi-step planning. 
The data is primarily sourced from the \textbf{Epic-Kitchens-100 \cite{damen2020epic}} using egocentric action annotations from \textbf{EgoPlan \cite{chen2023egoplan}}.
The resulting dataset pairs $(SG_t, \text{Instruction})$ with the target $SG_{goal}$, training the model to infer the action required to meet user goals. 
\textbf{After augmentation, this dataset encompasses 288k samples}.

\subsection{GSR Model Training} 
The GSR is based on the Qwen3-8B architecture \cite{qwen8B_system_card_2025}. 
We employ Grouped-Query Attention, where 32 query heads share 8 key value heads. 
Rotary Position Embeddings are used to improve long-context modeling, which is necessary for processing verbose scene-graph representations.
At inference, the GSR takes the scene graph and the user instruction, and outputs the next action command. 
Training is conducted in two separate stages: (i) Supervised Fine-Tuning (SFT) to activate structured reasoning, followed by (ii) Reinforcement Fine-Tuning (RFT) to align GSR with execution constraints.

\subsubsection{Training Stage-1 (SFT)}
The first training stage focuses on learning explicit scene understanding, goal interpretation, and action reasoning. 
We train on a large mixture of trajectories converted into instruction-following formats. 
We adopt parameter-efficient fine-tuning (PEFT) using Low-Rank Adaptation (LoRA) applied to all linear layers. 
The resulting supervised policy $\pi_{\mathrm{SFT}}$ is trained to produce structured intermediate reasoning followed by action commands.
However, we observe that models trained with SFT often exhibit reasoning artifacts, such as hallucinated object identifiers, thereby motivating the need for stage-2 training.

\begin{table}[ht]
\centering
\caption{Training Hyperparameters for GSR Models}
\label{tab:hyperparams}
\begin{tabular}{@{}lc@{}}
\toprule
\textbf{Hyperparameter} & \textbf{Qwen3-8B} \\ \midrule
Model Scale & 8.2 B \\
Non-Embedding Parameters & 6.95 B \\
Max Context Length & 131,072 (YaRN) \\
Training Sequence Length & 2,048 \\
Precision & bfloat16 \\
Training Method & LoRA (all-linear) \\
LoRA Rank ($r$) & 8 \\
LoRA Alpha ($\alpha$) & 32 \\ \midrule
Optimizer & AdamW \\
Learning Rate & $1 \times 10^{-4}$ \\
LR Scheduler & Cosine w/ Warmup \\
Warmup Ratio & 0.05 \\
Training Epochs & 20 \\
Per-Device Batch Size & 4 \\
Gradient Accumulation Steps & 16 \\
Global Batch Size & 128 \\ \bottomrule
\end{tabular}
\end{table}

\subsubsection{Training Stage-2 (RFT)}
To address the limitations of SFT and better align the model with physical execution constraints, we introduce a second stage of RFT using Group Relative Policy Optimization (GRPO). 
The optimization objective maximizes a task-specific reward function $R$ that targets three common failure modes in step-by-step embodied reasoning: (i) multi-step hallucinations, (ii) object grounding errors, and (iii) incorrect termination decisions. 
The total reward $R_{\text{total}}$ is defined as a weighted sum of these three components.
\begin{equation}
R_{total} = \lambda_1 R_{S} + \lambda_2 R_{G} + \lambda_3 R_{T}
\end{equation}

\paragraph{Step-wise Constraint Reward}
This term addresses the tendency of the SFT model to hallucinate future plans by generating multiple actions in a single output (e.g., “Pick A, then Place B”), whereas the execution engine requires strictly step-by-step action generation. We impose a strict format penalty: given the predicted action token sequence $A_{\text{pred}}$, the reward $R_{S}$ encourages the model to output exactly one atomic action tuple following the reasoning process.

\begin{equation}
R_{S} =
\begin{cases}
1,      & \text{if } a_{\text{pred}}=\texttt{END}\land \text{Satisfied}(S,G)\\[2pt]
-\alpha, & \text{if } N(A_{\text{pred}})>1\\[2pt]
0,      & \text{otherwise}
\end{cases}
\end{equation}
where $N(\cdot)$ counts the number of executable action primitives in the response. This forces the model to focus on the immediate next step rather than open-loop planning.

\begin{figure*}
    \centering
    \includegraphics[width=1.0\textwidth,height=0.26\textheight]{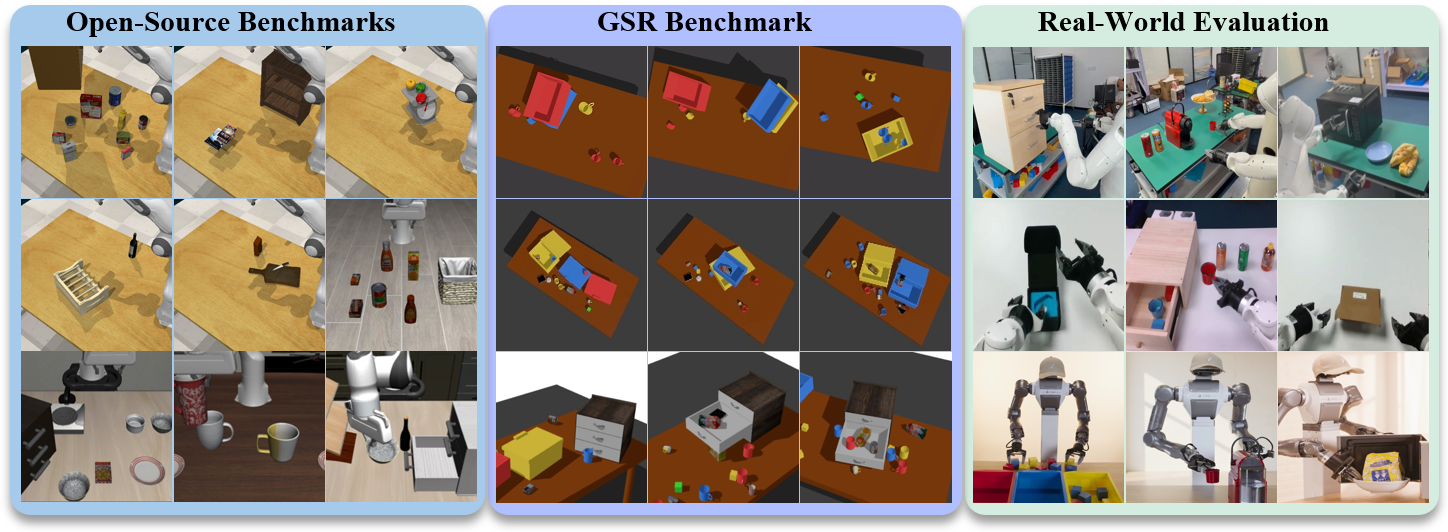}
    \caption{our evaluation comprises three components: (1) open-source benchmarks with 230 tasks from RLBench and LIBERO; (2) GSR task suites with 180 long-horizon tasks of varying difficulty targeting semantic object disambiguation, spatial-aware sequencing, and goal-conditioned generalization; and (3) real-world evaluations across diverse tasks.}
    \label{fig:placeholder}
\end{figure*}

\paragraph{Scene-Graph Grounding Reward}
This term targets object hallucination, a common failure mode in which the model generates objects that are not exist, such as typographical variants (e.g., “$apple$” instead of “$apple\_01$”). 
The reward enforces strict object grounding: given the predicted target object identifier $obj_{\text{pred}}$ and the set of valid node labels $V_{SG}$ in the current scene graph, the model is rewarded only when $obj_{\text{pred}} \in V_{SG}$.

\begin{equation}
R_{G} = \mathbb{I}(obj_{pred} \in V_{SG})
\end{equation}
where $\mathbb{I}(\cdot)$ is the indicator function. This hard constraint eliminates execution errors caused by invalid object references.

\paragraph{Termination Reward}
This term addresses termination errors, including premature termination (outputting “Task End” before the goal is achieved) or infinite looping (continuing to generate actions after task completion). 

\begin{equation}
R_{T} =
\begin{cases}
1,  & \text{if } a_{\text{pred}}=\texttt{END}\ \text{and}\ \text{Satisfied}(S_{\text{env}},G) \\[4pt]
-\beta, & \text{if } a_{\text{pred}}=\texttt{END}\ \text{and}\ \neg\,\text{Satisfied}(S_{\text{env}},G) \\[4pt]
0,  & \text{otherwise}
\end{cases}
\end{equation}
where $S_{\text{env}}$ denote the current environment state and $G$ the goal condition. The reward differentiates between correct termination and incorrect stopping behaviors.

\subsection{Real-world Robot Deployment} \label{sec: real_robot}
To deploy the GSR framework on a physical robot, we implement a meta-skill library that bridges high-level reasoning with low-level execution.
At the front end, given a action command, VFM computes a set of oriented 3D keypoints , which specify the start state, rational interaction points, and goal.
At the back end, we employ a behavior generation module based on geometric fabrics. 
Each behavior encoded by a geometric fabric follows the form:
\begin{equation}
    M_f(q_f, \Dot{q}_f) \Ddot{q}_f + f_f(q_f, \Dot{q}_f)+f_{\pi}(a)=0
\end{equation}
where \( q_f \) and \( \dot{q}_f \) denote the generalized coordinates and velocities in the fabric space (e.g., joint or end-effector space). 
The state-dependent metric \( M_f(q_f, \dot{q}_f) \) shapes the motion geometry by encoding task priorities, while \( f_f(q_f, \dot{q}_f) \) captures intrinsic fabric dynamics such as attraction, damping, and obstacle avoidance. 
The action-conditioned term \( f_{\pi}(a) \), specified by the selected action primitive \( a \), injects task intent (e.g., target, interaction force constraints, manipulability terms). 
Our physical system uses a Tianji Marvin-6 7-DoF dual-arm robot, with an Intel RealSense D435 RGB-D camera on the head.

\section{Experiments} \label{sec: exp}
We conduct a comprehensive evaluation of GSR, including one-shot reasoning under diverse task configs in \textbf{RLBench} and \textbf{LIBERO}, spatial-aware, goal-conditioned, and long-horizon reasoning in the \textbf{GSR-Bench}, and real-world robot implementation to assess performance, as shown in Figure \ref{fig:placeholder}.

\subsection{Evaluation on RLBench and LIBERO}
\begin{table}[!htbp]
\centering
\caption{RLBench LIBERO }
\label{tab:tableTab}
\begin{tabular}{lcc}
\toprule
Platform & Category & Amount \\
\midrule
\multirow{5}{*}{RLBench}
 & Kitchen Operations \textbf{(KO)}   & 25 \\
 & Pick and Place \textbf{(PP)} & 27 \\
 & Switch Operations for Containers \textbf{(SC)}   & 31 \\
 & Controller Operations \textbf{(CO)}       & 7 \\
 & Stacking and Assembly \textbf{(SA)}       & 10 \\  
 \cmidrule{1-3}
\multirow{4}{*}{LIBERO}
 & LIBERO\_Object   & 10 \\
 & LIBERO\_Spatial & 10 \\
 & LIBERO\_Goal       & 10 \\
 & LIBERO\_Long       & 10 \\
 & LIBERO\_Short       & 90 \\
\bottomrule
\end{tabular}
\end{table}

\begin{table*}[h!]
    \centering
    \caption{Performance comparison on the RLBench and LIBERO Benchmark.}
    \small  
    
    \label{tab:comparison_on_the_RLBench_libero}
    \begin{tabular}{lccccc|ccccc}  
    \toprule
    \multirow{2}{*}{\textbf{Method}} & \multicolumn{5}{c}{\textbf{RLBench}} & \multicolumn{5}{c}{\textbf{LIBERO}} \\ \cmidrule(lr){2-6} \cmidrule(lr){7-11} 
                     & KO  & PP & SC & CO & SA & Object  & Spatial & Goal & Short & Long  \\
    \midrule
    GPT-5            & 80.00 & 58.20 & 45.00 & 33.30 & 15.00 & 81.00  & 79.00 & 69.00 & 80.00 & 57.00    \\
    
    Gemini-2.5-pro   & \textbf{95.00} & \textbf{85.60} & 51.50 & 46.70 & 30.00 &\textbf{87.00}  & 86.00 & \textbf{74.00} & 86.00 & \textbf{68.00}    \\
    
    DeepSeek-V3      & 90.00 & 70.00 & 64.30 & 42.00 & 32.50 & 82.00  & 77.00 & 73.00 & 80.00 & 61.00   \\
    
    Claude-sonnet-4-5 & 30.00 & 51.00 & 63.20 & 36.70 & 20.00 & 82.00  & \textbf{88.00} & 74.00 & 83.00 & 62.00  \\

    Qwen3-8B         & 25.00 & 51.00 & 56.40 & 33.30 & {26.70} & 84.00  & 81.00 & 72.00  & 82.00 & 54.00    \\
    
    \cellcolor{gray!20}GSR-8B (Ours) &
    \cellcolor{gray!20}92.50 &
    \cellcolor{gray!20}82.00 &
    \cellcolor{gray!20}\textbf{68.50} &
    \cellcolor{gray!20}\textbf{52.50} &
    \cellcolor{gray!20}\textbf{41.20} &
    \cellcolor{gray!20}86.50 &
    \cellcolor{gray!20}\textbf{89.20} &
    \cellcolor{gray!20}81.00 &
    \cellcolor{gray!20}85.00 &
    \cellcolor{gray!20}\textbf{82.40}\\
    
    \bottomrule
    \end{tabular}   
\end{table*}

\begin{figure}[htbp]
  \centering
  \includegraphics[width=1\columnwidth]{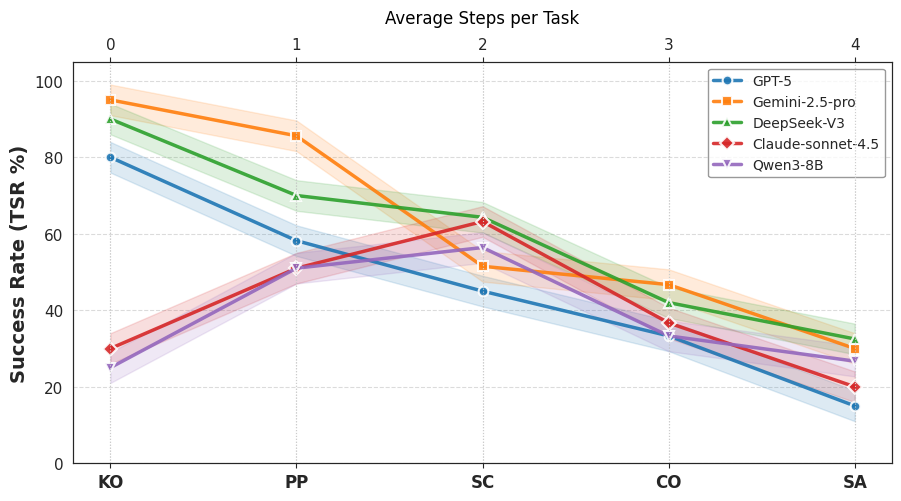}
  \caption{Comparison of model performance on RLBench, showing success rates with respect to task complexity.} 
  \label{fig:rlbench-comparsion-01}
\end{figure}

This experiment evaluates the model’s \textbf{general reasoning capability} in \textbf{manipulation tasks without task-specific training}.
The model must generate the \textbf{command} conditioned on given \textbf{initial state}, infer the underlying \textbf{goal}, and predicts the correct \textbf{actions}. 
We select five LLMs as baselines, including \texttt{GPT-5} \cite{openai_gpt5_system_card_2025}, \texttt{Gemini-2.5-Pro} \cite{comanici2025gemini}, \texttt{DeepSeek-V3} \cite{deepseekai2024deepseekv3technicalreport}, \texttt{Qwen3-8B} \cite{qwen8B_system_card_2025}, and \texttt{Claude-Sonnet-4.5} \cite{anthropic_claude_sonnet4.5_model_card_2025}.
To deploy LLMs in manipulation tasks, we integrate a scene-graph–based perception module with an LLM control interface (see Appendix~\ref{app:simulation evalution}), which maps language commands to corresponding low-level actions.
We evaluate GSR in a zero-shot setting on RLBench and LIBERO (Table \ref{tab:tableTab}) without any fine-tuning. For the other models, we design task-specific prompts for each LLM to ensure their performance is well optimized during evaluation. 
On \textbf{RLBench}, we evaluate 100 tasks across five representative categories: \textbf{Kitchen Operations (KO)}, \textbf{Pick and Place (PP)}, \textbf{Switch Operations for Containers (SC)}, \textbf{Controller Operations (CO)}, and \textbf{Stacking and Assembly (SA) (see Appendix~\ref{app:simulation evalution})}.
On \textbf{LIBERO}, we evaluate 100 tasks across four subsets: \textbf{LIBERO-Spatial}, \textbf{LIBERO-Object}, \textbf{LIBERO-Goal}, and \textbf{LIBERO-Long}.
We run ten trials per task and report the \textbf{success rate}. 
Detailed results are provided in Table~\ref{tab:comparison_on_the_RLBench_libero} and Figure \ref{fig:rlbench-comparsion-01}.

\begin{figure*}
    \centering
    \includegraphics[width=1.0\linewidth]{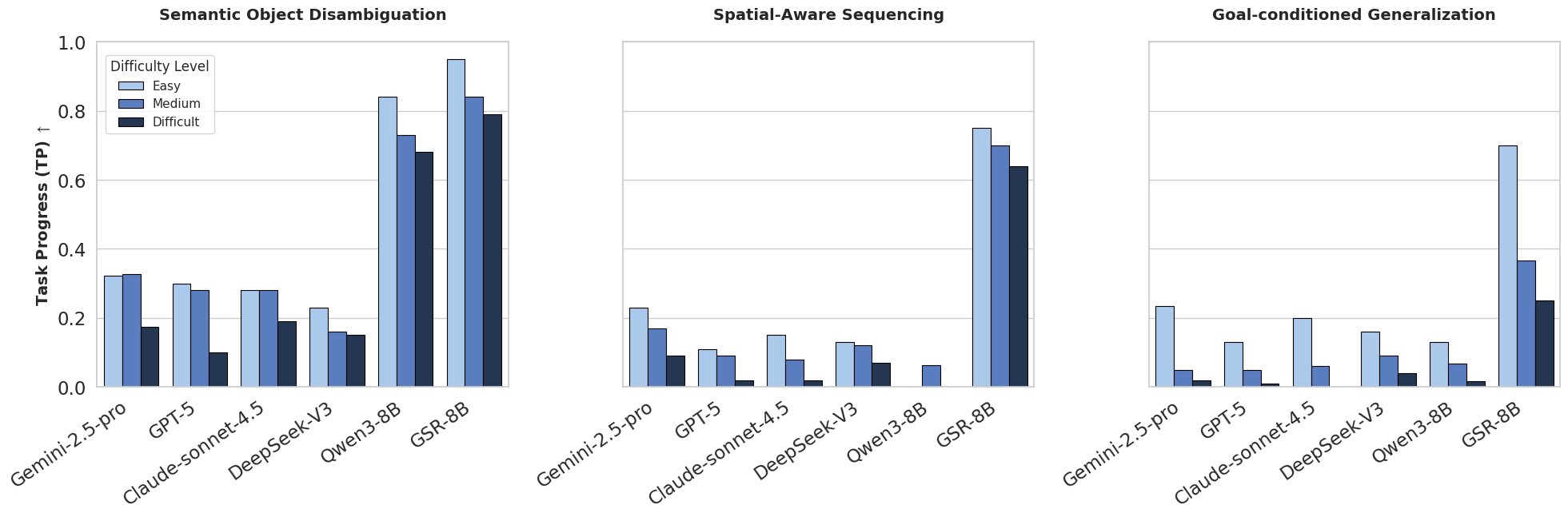}
    \caption{Evaluation on GSR-Bench via Task Progress (TP \%), demonstrates that GSR achieves SOTA performance across all three evaluation aspects, with strong advantages on long-horizon tasks involving complex spatial constraints and goal reasoning.}
    \label{fig:TP result}
\end{figure*}

From the results, several consistent trends emerge.
Across all models, performance degrades as tasks involve more complex spatial reasoning and longer action sequences. Simple pick-and-place tasks impose limited spatial constraints, whereas stacking and assembly require reasoning over object relations and intermediate states, resulting in sharp performance drops (e.g., \texttt{Gemini-2.5-Pro} decreases from 95\% on PP to 30\% on SA).
This degradation intensifies with increasing task horizons, as longer sequences amplify reasoning errors and hinder recovery; for instance, \texttt{DeepSeek-V3} drops from 90\% to 32.5\% as the average number of steps increases from 1.85 to 3.9.
A similar pattern is observed on LIBERO, where longer-horizon and goal-conditioned tasks (LIBERO-Long and LIBERO-Goal) consistently yield lower success rates across all models.

\subsection{Evaluation on GSR-bench}
\textbf{GSR-Bench} evaluates long-horizon task reasoning under both spatial and semantic constraints. 
It consists of 180 tasks, with an average planning horizon exceeding 10 steps. 
Our evaluation focuses on three aspects: 
\textbf{(1) Semantic Object Disambiguation (SOD)}, which assesses reasoning under varying object semantics; 
\textbf{(2) Spatial-Aware Sequencing (SAS)}, which evaluates reasoning over physical causality and spatial constraints; 
and \textbf{(3) Goal-conditioned Generalization (GCG)}, which measures reasoning across diverse abstract goals. 
For each aspect, we define three difficulty levels: \textit{easy}, \textit{medium}, and \textit{difficult} (see Appendix~\ref{Appendix:GSR}). 
We use \textbf{Task Progress (TP)} as the evaluation metric, and the experimental results are reported in Fig.~\ref{fig:TP result}.

\begin{table}[hp]
\centering
\caption{Analysis of GSR to Scene Graph Noise.}
\label{tab:sg_noise}

\begin{tabular}{m{2.0cm}@{\hspace{2pt}}c|@{\hspace{2pt}}ccc}
\toprule
Task Type & Noise & Easy & Med.\ & Hard \\
\midrule
\multirow{3}{*}{\centering Task-SOD}
 & 0\%  & 1.000 & 0.980 & 0.969 \\
 & 5\%  & 0.960 & 0.940 & 0.910 \\ 
 & 10\% & 0.900 & 0.880 & 0.850 \\
\midrule
\multirow{3}{*}{\centering Task-SAS}
 & 0\%  & 0.975 & 0.980 & 0.940 \\
 & 5\%  & 0.920 & 0.910 & 0.860 \\
 & 10\% & 0.850 & 0.830 & 0.750 \\
\midrule
\multirow{3}{*}{\centering Task-GCG}
 & 0\%  & 0.917 & 0.367 & 0.250 \\
 & 5\%  & 0.800 & 0.300 & 0.180 \\
 & 10\% & 0.680 & 0.220 & 0.120 \\
\bottomrule
\end{tabular}
\end{table} 

Across all evaluated tasks, \textbf{TP} consistently degrades as task difficulty increases from \textit{Easy} to \textit{Difficult}. 
For instance, \texttt{GPT-5}'s TP on \textit{Spatial-Aware Sequencing} drops sharply from 0.11 (Easy) to 0.02 (Difficult). 
This trend is consistent across models and task categories: as object density, spatial occlusion, and instruction ambiguity increase, embodied reasoning performance degrades.
Similar performance drops are observed on RLBench and LIBERO as tasks require more complex spatial reasoning and longer action sequences.
The effect is most pronounced in long-horizon tasks, where prompting-based LLMs fail to consistently maintain spatial and goal constraints across steps.

Among all tasks, \textit{goal-conditioned tasks} are the most challenging. 
This is mainly for two reasons.
First, the instructions are often more abstract and vague (for example, “sort objects of different colors into different drawers”), which requires stronger language understanding and goal interpretation.
Second, \textit{goal-conditioned tasks} usually involve more complex spatial constraints and longer action sequences, which demand multi-step planning and reasoning over intermediate states.
The quantitative results clearly reflect this trend. 
For example, \texttt{Gemini-2.5-Pro} achieves TP of $0.259$, $0.163$, and $0.101$ on Object, Spatial, and Goal tasks, respectively. \texttt{GPT-5} shows a similar pattern, with TP values of $0.227$, $0.073$, and $0.063$.
Interestingly, some models perform slightly better on \textit{Goal–Easy tasks} than on \textit{Spatial–Easy tasks}. For instance, \texttt{Claude-Sonnet 4.5} achieves a TP of $15\%$ on \textit{Spatial–Easy} but improves to $20\%$ on \textit{Goal–Easy}. 
This occurs because \textit{Goal–Easy} tasks exhibit lower spatial complexity: some cases do not involve  operations on stacked containers or articulated objects.
Overall, \textit{goal-conditioned tasks} are the most challenging, as they require interpreting abstract instructions alongside spatial reasoning, where GSR consistently outperforms prompting-based baselines.

\subsection{Analysis of Robustness to Perceptual Noise}
To evaluate robustness to perception errors, we perturb input scene graphs at inference time on GSR-8B without retraining, as shown in Table \ref{tab:sg_noise}. 
In GSR-Bench, we randomly drop or flip scene-graph relations with noise ratios of 5\% and 10\%, and evaluate performance using Task Progress (TP).
GSR remains robust under moderate noise: with 5\% perturbation, TP decreases only slightly across tasks (e.g., Semantic Object Disambiguation drops from 1.00 / 0.98 / 0.97 to 0.96 / 0.94 / 0.91 for Easy/Medium/Hard, respectively). At 10\% noise, performance degrades further but remains stable overall.
Sensitivity varies by task type: Goal-conditioned Generalization is most affected (Hard TP drops from 0.25 to 0.12), while Semantic Object Disambiguation and Spatial-Aware Sequencing maintain higher performance under the same noise levels.

\subsection{Demonstration in Real-World Tasks}\label{section: realworld evaluation}
This experiment evaluates GSR in three real-world settings, each targeting a distinct aspect of its reasoning and generalization capabilities. 
First, we assess GSR on general pick-and-place tasks involving common objects, evaluating its reasoning ability to generate appropriate behaviors in response to natural language commands. 
Second, we evaluate GSR on long-horizon sorting tasks as in GSR-Bench, examining its performance across diverse spatial configurations and goal conditions that require extended planning. 
Third, we demonstrate GSR on four daily-life tasks, detailed results are shown in the supplementary videos.

\section{Conclusion} \label{sec: conclusion}
In this study, we presented GSR, an embodied reasoning framework that operates over explicit, structured world representations. By reasoning on scene graphs encoding object states and relations, rather than direct perception-to-action mappings, GSR enables transferable physical commonsense reasoning and improved generalization in manipulation. Extensive evaluations in simulation and on physical robots show that GSR achieves strong zero-shot performance, long-horizon reasoning, and robustness to real-world perturbations.

\noindent\textbf{Limitations.} 
our work has several limitations. 
First, the current framework relies on predefined action primitives, which constrains flexible action generation and adaptation between successive steps in long-horizon tasks. 
Second, the vision representations used to construct scene graphs remain limited in their ability to capture fine-grained geometry, subtle object states, and nuanced physical properties, which can affect reasoning accuracy in visually complex or cluttered environments.

\noindent\textbf{Impact.}
The findings presented in this work indicate that grounding decision-making in explicit world representations supports more systematic generalization, sustained reasoning over time, and adaptive behavior in open-world embodied settings.
Future work will explore more expressive and adaptive action policy, as well as improved perception models that yield more precise world representations to support finer-grained reasoning and control.

\bibliographystyle{IEEEtran}
\bibliography{bibtex/IEEE_available}

\clearpage
\appendix

\section{Appendix}
\subsection{Details of Vision Foundation Model} 
\label{Appendix: VFM}
This section provides details of the open-world visual foundation model that we use. 
To accurately represent an object's comprehensive information, we design an interpretable visual information extraction framework. 
By analyzing task objectives, we retrieve initialization parameters for the model and set up corresponding information extraction workflows.
After initialization, we use YOLO-E  for efficient open-world object detection and segmentation. 
To extract object keypoint information, we fine-tuned the detection head, which enables us to extract keypoint data for specific objects. 
To continuously track points, we simplified Spatial-TrackerV2 and optimize the feature extraction module. 
This ensures the point tracker maintains stable tracking during robotic operations or movement while keep a fixed frequency with the overall model.

We also consider that real-time updates to the scene graph require rapid acquisition of objects' 3D spatial coordinates with high-frequency updates. We develop a lightweight 6D object pose estimation model based on Gen-pose++.

\subsubsection{Detection Details for Different Objects}
Our detection model design is based on the three object types described in the experimental section.

\textbf{Rigid Objects:} For rigid objects, such as basic geometric shapes (cubes), no special functional design is necessary. 
We employ only basic 6D pose detection to provide GSR with stable spatial detection results, supporting accelerated scene graph construction. 
For the colored cube sorting task, we only require accurate detection to precisely identify the positions of different colored cubes and their target placement areas. 
This enables us to obtain the spatial relations of all objects and compute the scene graph, which is ultimately handed over to the GSR for operation planning. 
To effectively represent dynamic scene relations, the detection model must maintain a fixed detection frame rate (15 Hz). 
Therefore, we developed a higher-frequency detection method to ensure GSR processes decisions based on the most up-to-date spatial relation information.

\textbf{Articulated Objects:} For fine manipulation tasks with articulated objects, we introduce keypoint detection to guide GSR for accurate planning.
Specifically, when a robot needs to operate a coffee machine, relying only on object-level 6D pose estimation or state-of-the-art visual language models (VLMs) is often insufficient for accurately locating key operable components on the machine (such as buttons and switches).
To address this, we propose a fusion representation method that combines detected component keypoint information in an explicit encoded form with the object's 6D pose estimation results. 
This integrated approach is unified within the scene graph computation framework. Experiments demonstrate that this fused information representation provides more stable and accurate geometric and semantic constraints for subsequent operation planning, significantly enhancing the system's performance in complex interactive tasks.

\textbf{Soft objects:} Traditional 6D pose estimation methods based on rigid body assumptions often prove ineffective when handling soft objects.
To address this, we employ a keypoint-tracking strategy that detects and tracks semantic keypoints on the object's surface in real time, providing frame-by-frame geometric state updates for manipulation.
This approach adapts to object deformation and pose changes while continuously outputting stable keypoint information and supporting real-time maintenance and updates of dynamic scene graphs.
In the pencil case opening/closing task, this method demonstrates strong robustness. Specifically, the system reliably detects the zipper position and its open/closed state, estimates the opening degree of the case in real time, guides the robotic arm to place writing instruments, and continuously tracks zipper movement until closure.
Experiments demonstrate that this keypoint tracking framework effectively overcomes operational challenges posed by geometric uncertainties in soft objects, maintaining task accuracy while preserving the consistency of scene understanding.

\subsubsection{Implementation of Real Scene Graph}
\begin{figure}
    \centering
    \includegraphics[width=0.9\linewidth]{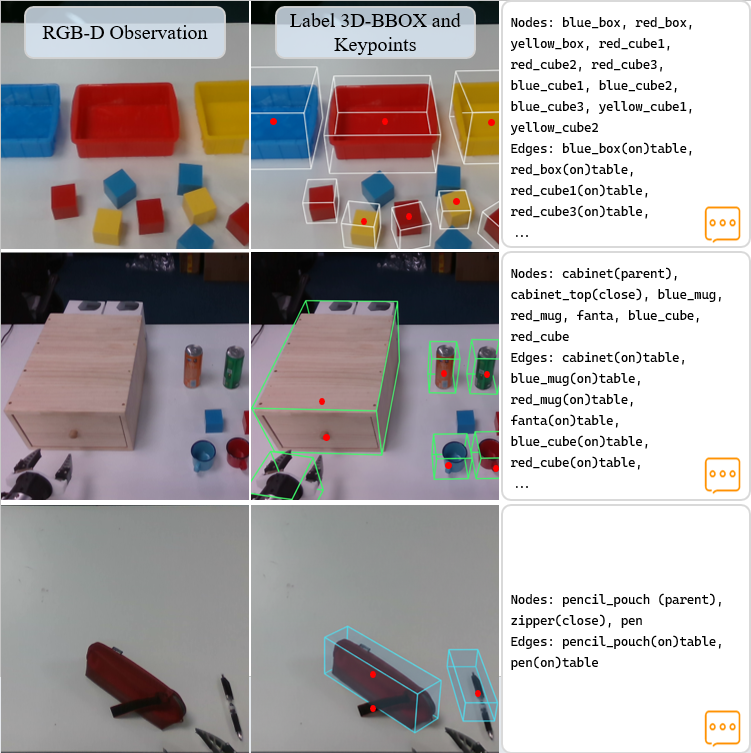}
    \caption{Results of Vision Model Detection and Real Scene Graph Generation}
    \label{fig:appendix_sgg}
\end{figure}
We tested the dynamic scene graph generation capability of our visual fusion model across three real-world scenes. 
Results in Figure\ref{fig:appendix_sgg} demonstrate that the model consistently provides the necessary information for scene graph generation. 
By explicitly incorporating object keypoint state information, we effectively enhance the LLM's understanding of the geometric and functional states of fine-grained components within images.
Regarding the specifics of visual information from real-world scenes to scene graph generation, as the core work of this paper does not address scene graph generation, we will perform detailed experimental data comparisons in future work.
\label{Appendix A}

\clearpage
\subsection{Manip-Cognition Data Engine Details} \label{app:data_details}
The Manip-Cognition dataset serves as the primary training source for the Grounded Scene-graph Reasoning (GSR) framework. It bridges the gap between raw visual demonstrations and structured symbolic reasoning. In the following sections, we provide details regarding Data Selection and Sources (Sec.~\ref{app:Data Selection}), Data Cleaning and Structural Normalization (Sec.~\ref{app:cleaning}), the Annotation Pipeline (Sec.~\ref{app:annotation}), and our Data Augmentation strategies (Sec.~\ref{app:Data Augmentation}).

\subsubsection{Data Selection and Sources} \label{app:Data Selection}

To cover diverse task structures and physical interaction types, we collected our data from 4 open-source embodied datasets and benchmarks.

\begin{itemize} 
\item \textbf{Behavior-1K \cite{li2024behavior} and Enact \cite{wang2025enact}}: We utilize 29 task scenarios from Behavior-1K, which provide full demonstration videos and scene graph ground-truth. By leveraging keyframe annotations from the Enact benchmark, we extract \textbf{856 unique Scene Graphs (SGs)} representing distinct action boundaries. After cleaning and augmentation, these form the foundational \textbf{Scene Grounding Data} (12,000 pairs). 
\item \textbf{EPIC-KITCHENS-100 \cite{damen2020epic} and Ego-Plan \cite{chen2023egoplan}}: To scale long-horizon reasoning, we process 614 long-form videos across 42 kitchen scenes. From the 9,138 unique tasks and 69,253 action keyframes annotated in Ego-Plan, we perform strict filtering to retain \textbf{6,000 complete episodic trajectories}, averaging 6 sub-steps per task. These trajectories are decomposed into specific reasoning modalities:
\begin{itemize}
\item \textbf{World Modeling}: Pairs of consecutive scene graphs $(SG_t, SG_{t+1})$ and their intervening action $A_t$ are used to predict edge variations: $(SG_t, A_t) \rightarrow \Delta SG_{\text{edge}}$.
\item \textbf{Forward Action Reasoning}: Sequences spanning multiple steps are used to bridge the gap between states: $(SG_t, SG_{t+n}) \rightarrow \{A_i\}_{i=t}^{t+n-1}$.
\item \textbf{Goal-Conditioned Planning}: Current scene graphs are paired with high-level task goals $G$ to predict the immediate next action: $(SG_t, G) \rightarrow A_t$.
\item \textbf{Goal Interpretation}: The initial state $SG_{\text{start}}$ and the final task-completion state $SG_{\text{goal}}$ are paired with the global instruction to train "goal dreaming": $(SG_t, \text{Goal Instruction}) \rightarrow SG_{\text{goal}}$.
\end{itemize}
\end{itemize}

\subsubsection{Data Cleaning and Structural Normalization} \label{app:cleaning}

To ensure structural reasoning is reliable and to minimise model hallucination, we implemented targeted cleaning procedures based on the specific characteristics of each data source.

\begin{itemize} 
\item \textbf{Behavior-1K Scene Graphs}: Raw SGs extracted from the Behavior-1K exhibited two main structural issues:
(1) Inconsistent relational syntax: Spatial relations were represented inconsistently, with redundant bidirectional labels sometimes being used (e.g. both '$A(ontop)B$' and '$B(under)A$'), and at other times only a single unidirectional predicate was used. This lack of continuity in expression creates conflicting supervision signals.
(2) Node-edge discrepancy: There were frequent mismatches between the objects listed in the SG nodes and those appearing in the edge predicates. If left uncorrected, these issues can lead to training instability and severe model hallucinations, such as the generation of false nodes or edges that are logically impossible. To address this, we first normalised all relations into a strictly unidirectional 'ontop' format. We then reconstructed the object node lists for every scene graph directly from the validated edge predicates.

\item \textbf{Ego-Plan Task Completion}: The original Ego-Plan labels often contained broken action chains that showed only the first few steps of a longer task. We used GPT-5 to check every video trajectory and removed any clip that did not reach a clear, functional end. The specific prompt used is shown in Figure \ref{fig:filtering}.

\begin{figure}
    \centering
    \includegraphics[width=0.9\linewidth]{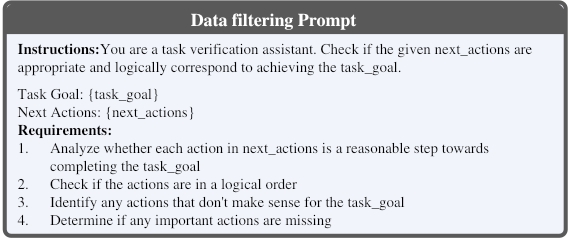}
    \caption{\textbf{Illustration of the Data Cleaning Prompt}. This prompt is designed to leverage GPT-5’s advanced reasoning capabilities for automated dataset refinement and noise reduction.}
    \label{fig:filtering}
\end{figure}

\end{itemize}

\subsubsection{The Data Annotation Pipeline} \label{app:annotation}
We developed a tiered annotation strategy to transform raw visual observations and scene graphs into the structured reasoning pairs required for the GSR framework.

\begin{figure*}
    \centering
    \includegraphics[width=0.9\linewidth]{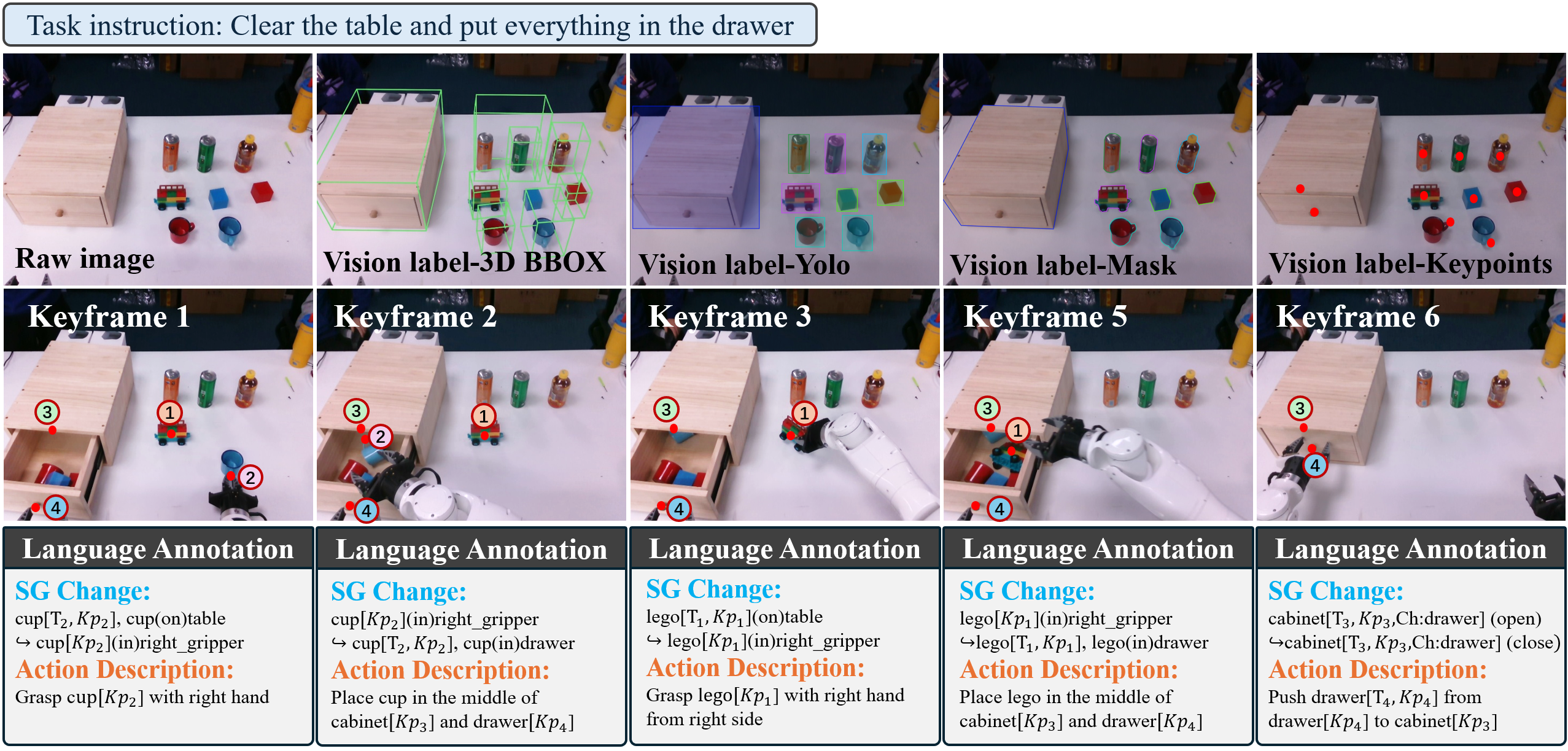}
    \caption{Annotation examples for data collected in real-world scenarios. $T_j$ denotes the 3D pose of the object, $Kp$ represents the semantic keypoints on the object, and $Ch$ indicates the child nodes.}
    \label{fig:Annotation real-world scenarios}
\end{figure*}

\begin{figure}
    \centering
    \includegraphics[width=0.9\linewidth]{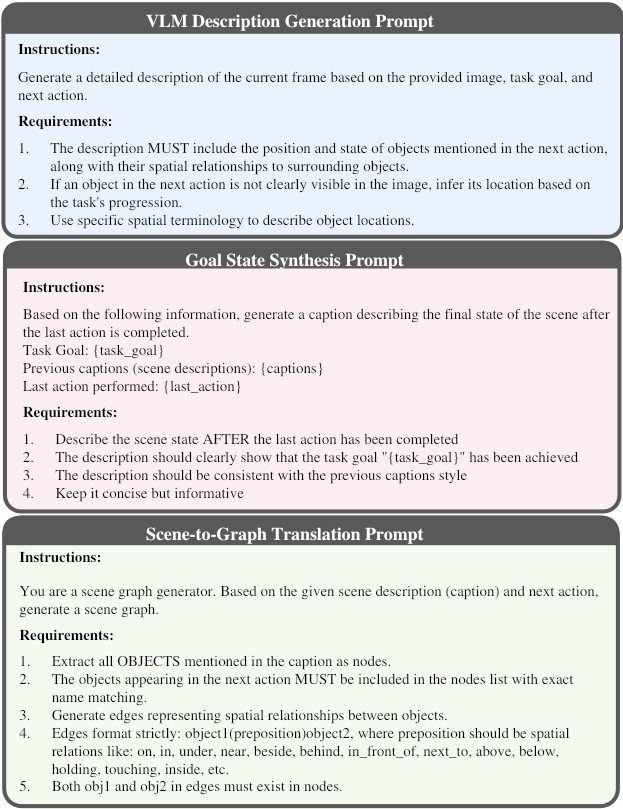}
    \caption{\textbf{Prompting paradigms for Episodic and Goal State Annotation}. The framework facilitates the synthesis of complete reasoning paths, ensuring logical consistency between environmental observations and final task objectives.}
    \label{fig:caption}
\end{figure}

\begin{itemize}
\item \textbf{Scene-to-Graph Grounding Annotation}:
To construct the \textit{Scene Grounding Data}, we utilized the cleaned SGs from Behavior-1K and Enact. Since these sources lack diverse natural language, we employed GPT-5 to generate distinct, linguistically varied description for each unique SG instance. These descriptions strictly preserve object identities and spatial relations while providing the stylistic diversity needed for robust visual-to-symbolic grounding.

\item \textbf{Episodic and Goal State Annotation}:
To transform visual keyframes from EPIC-KITCHENS into complete reasoning trajectories, we implemented a three-stage pipeline, the prompt used in each stage is shown in the Figure \ref{fig:caption}:
\begin{itemize}
    \item \textbf{Stage 1: VLM Description Generation}: A locally deployed Qwen3-VL-8B generates detailed natural language descriptions of the visual state for each action boundary keyframe.
    \item \textbf{Stage 2: Goal State Synthesis}: Since Ego-Plan annotations do not label the true task completion state. We utilized GPT-5 to reason over the task instruction and preceding action descriptions to synthesize a structured description of the intended final state (e.g., the configuration of a "folded box").
    \item \textbf{Stage 3: Scene-to-Graph Translation}: All generated descriptions (from both keyframes and synthesized goal states) are processed by a specialized LLM (fine-tuned on the Enact SG-Description pairs) to translate the text into standardized Scene Graphs ($SG_t$ and $SG_{goal}$).
\begin{table*}[h]
\centering
\caption{Summary of Data Augmentation Factors and Final Dataset Scale.}
\label{tab:augmentation_summary}
\small
\begin{tabular}{@{}lcccc@{}}
\toprule
\textbf{Data Modality} & \textbf{Base Size} & \textbf{Augmentation Methods} & \textbf{Multiplier} & \textbf{Final Quantity} \\ \midrule
\textbf{Scene Grounding} & 856 & Paraphrasing (3) $\times$ Shuffling (3) $\times$ Swapping (2) & 18$\times$ & $\approx$ 15,000 \\
\textbf{Planning Data} & 36,000 & Shuffling (3) $\times$ Swapping (2) $\times$ Rephrasing (2) $\times$ 3 Tasks & 36$\times$ & 1,296,000 \\
\textbf{Goal Interpretation} & 6,000 & Shuffling (3) $\times$ Swapping (2) $\times$ Rephrasing (2) $\times$ End-state (4) & 48$\times$ & 288,000 \\ \bottomrule
\end{tabular}
\end{table*}
\item \textbf{Real-World Evaluation Data Annotation}:
For the physical robot experiments described in Section~\ref{section: realworld evaluation}, we recorded synchronized RGB video streams and robot internal states. During data collection, action boundary keyframes were actively triggered and logged to isolate specific sub-task segments. The annotation for this evaluation set is divided into two granular layers:
\begin{itemize}
\item \textbf{High-Fidelity Visual Annotation}: We sampled the raw image stream at a 1/10 decimation rate. Each sampled frame was manually annotated with object-centric visual metadata, including 3D poses ($T_j$), YOLO bounding boxes, segmentation masks, and semantic keypoints ($K\!p$).
\item \textbf{Fine-Grained Linguistic and Symbolic Annotation}: For each keyframe, we annotated the natural language action description and the corresponding scene graph variations. Leveraging the dense visual metadata, these scene graphs utilize a higher level of granularity than our large-scale training set. Specifically, each object node is represented as a tuple:$o_j = [T_j, \mathcal{K}\!p_{1..n}, \mathcal{C}\!h_{1..m}]$.

where $T_j$ denotes the object's 3D pose, $\mathcal{K}\!p$ represents the semantic keypoints, and $\mathcal{C}\!h$ indicates the hierarchical child nodes. This detailed representation allows for precise action grounding, such as "Push drawer[$T_4, K\!p_4$] from drawer[$K\!p_4$] to cabinet[$K\!p_3$]", enabling a more rigorous evaluation of the agent's spatial reasoning capabilities in real-world scenarios (see Figure ~\ref{fig:Annotation real-world scenarios}).\end{itemize}

\end{itemize}
\end{itemize}

\subsubsection{Data Augmentation and Final Dataset Composition} \label{app:Data Augmentation}
\begin{figure}
    \centering
    \includegraphics[width=0.9\linewidth]{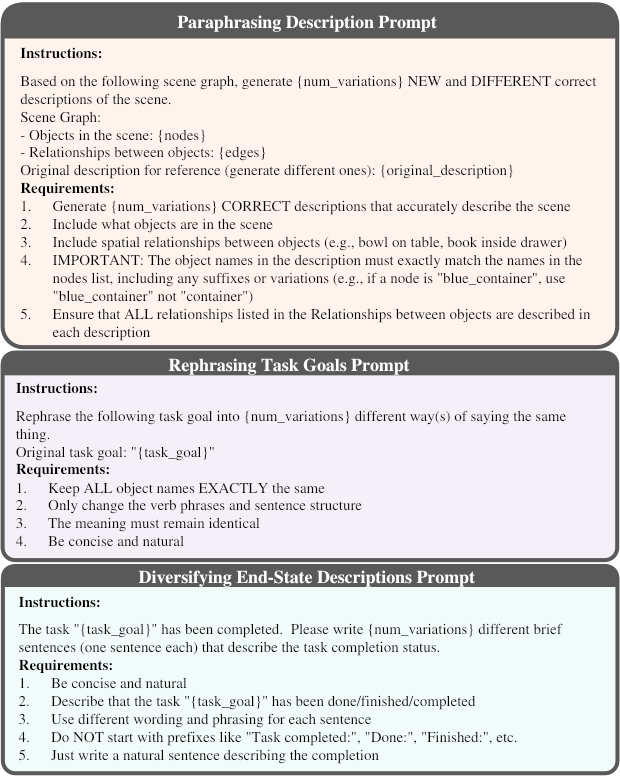}
    \caption{\textbf{Taxonomy of Data Augmentation Prompts}. The framework comprises three hierarchical strategies: Paraphrasing Descriptions for scene grounding enrichment, Task Goal Rephrasing, and End-State Diversification to facilitate data scaling within planning and goal interpretation tasks.}
    \label{fig:augment}
\end{figure}

To enhance the robustness of the GSR framework against linguistic variability and structural noise, we applied systematic data augmentation across all data modalities. And the final scale and composition of the Manip-Cognition dataset are also detailed below.
\begin{itemize}
\item \textbf{Scene Grounding Data Augmentation}:
Starting with the base SG-description pairs from Behavior-1K and Enact, we applied 3 types of expansion:
\begin{itemize}
    \item \textbf{Paraphrasing Description ($\times 3$)}: Each description was rephrased twice using GPT-5, ensuring identical semantics and consistent object identities, the prompt is shown in Figure \ref{fig:augment}.
    \item \textbf{Shuffling Graph Order ($\times 3$)}: To ensure the model is invariant to input order, the sequence of nodes and edges within the SG string was randomly reordered.
    \item \textbf{Replacing Object Names ($\times 2$)}:  We used a list of alternatives to replace common object names (e.g. 'cup' with 'mug') consistently throughout the SG and its description.
This resulted in a total of approximately \textbf{15,000} grounding pairs.
\end{itemize}

\item \textbf{Planning and Goal Interpretation Data Scaling}:
Based on the 6,000 cleaned episodic trajectories from Ego-Plan (averaging 6 steps per task, totaling 36,000 state-action pairs), we implemented a multi-factor augmentation strategy:
    \begin{itemize} 
    \item \textbf{Shuffling Graph Order  ($\times 3$)}: We randomly reordered the sequence of objects and relations within the scene graph text without changing the actual spatial relations (e.g. changing 'A on B, B on C' to 'B on C, A on B'). This teaches the model to focus on the underlying structure rather than the order of the input.
    \item \textbf{Replacing Object Names  ($\times 2$)}: We used a list of alternatives to replace common object names (e.g. 'cup' with 'mug') consistently throughout the SG and its description. This ensures the model understands the functional category of an object rather than just memorizing a specific label. 
    \item \textbf{Rephrasing Task Goals  ($\times 2$)}: We used GPT-5 to rewrite the original task instructions in different styles (e.g., changing "put the book in the bag" to "place the textbook inside the backpack"), the prompt is shown in Figure \ref{fig:augment}. This helps the model handle various ways humans might give commands. 
    \item \textbf{Diversifying End-State Descriptions  ($\times 4$) }: For the goal interpretation tasks, we generated four different textual descriptions for the same final scene, using GPT-5 with prompt shown in Figure \ref{fig:augment}. This helps the model more reliably "imagine" what a successful task completion looks like.
    \end{itemize}
\end{itemize}

Through the systematic selection, cleaning, annotation, and augmentation procedures described above, the Manip-Cognition dataset has been scaled to provide dense supervision across all reasoning modalities. The final composition and the specific impact of each augmentation factor are summarized in Table~\ref{tab:augmentation_summary}.

\clearpage
\subsection{SIMULATION EVALUATION DETAILS} \label{app:simulation evalution}
\begin{figure}[htbp]
  \centering
  \includegraphics[width=0.95\columnwidth]{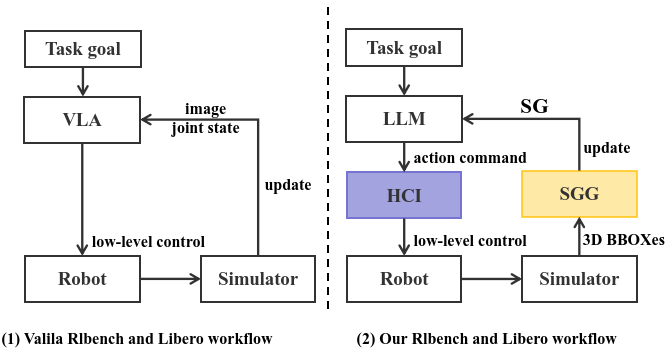}
  \caption{Architectural comparison of the standard simulation environments and our proposed framework.} 
  \label{fig:rlbench-sh-plan}
\end{figure}

The Vanilla RLBench and LIBERO systems execute tasks using image and joint state information, interacting directly with the simulator through low-level control commands. To integrate LLMs into RLBench and LIBERO and evaluate their task performance across diverse scenarios, as shown in Figure \ref{fig:rlbench-sh-plan}, we enhance the existing systems by incorporating HCI (High-Level Control Interface, see \ref{subsubsection:HCI}) and SGG (Scene Graph Generator, see \ref{subsubsection:SSG}). These enhancements enable the system to effectively control the robot based on action commands generated by the model while leveraging scene graphs to represent changes in environmental state.

\begin{table}[!htbp]
\centering
\caption{RLBench and LIBERO Status}
\label{tab:obj state}
\begin{tabular}{>{\centering\arraybackslash}m{2.2cm}|>{\centering\arraybackslash}m{0.3\linewidth} |>
{\centering\arraybackslash}m{0.3\linewidth}}
\toprule
\textbf{State Type} & \textbf{RLBench Status} & \textbf{LIBERO Status} \\
\midrule
Robot State & Holding/Empty & Holding/Empty  \\ 
\midrule
Spatial Relation  & Under/OnTop/Beside , Inside/Contains & On/Up/In/Stack \\
\midrule
Object State & Open/Close, On/Off, Empty/Full, Folded/Unfolded & Open/Close, Turn on/Turn off\\
\bottomrule
\end{tabular}
\end{table}

\begin{table}[!htbp]
\centering
\caption{RLBench and LIBERO Objects}
\label{tab:RLBench Objects}
\begin{tabular}{>{\centering\arraybackslash}m{2.8cm}|>{\centering\arraybackslash}m{0.58\linewidth}}
\toprule
\textbf{Category} & \textbf{Objects} \\
\midrule
Object Manipulation and Placement
  & Cuboid, Plane, Sphere, cylinder, cube, triangular prism, book, bottle, cap, chicken, chocolate jello, coffee, computer, crackers, cup, dish, dollar stack, jar, jar lid, knife, laptop holder, lid, moon, mustard, phone, phone case, plant, plate, plug, rubbish, saucepan, saucepan lid, scoop with spatula spatula, shoe, soup, spam, star, steak, stick, strawberry jello, sugar, tomato, tuna, tv remote, umbrella, usb, wine bottle, Table, base, bottom, drawer legs, item, left board, left frame, left handle, stand \\
\midrule
Container and Switch Operations
  & bin, bookshelf, box base, box lid, cabinet, cabinet base, cupboard, dish rack, door, door frame, door handle, drawer, drawer frame, fridge base, grill, knife block, microwave, microwave door, oven, oven door, plate stand, rack, safe, safe door, shelf, switch, tap, toilet, toilet seat up seat, tv, washer, window \\
\midrule
Stacking and Assembly Tasks
  & basket, ketchup, block pyramid block, block pyramid distractor block, jenga cuboid, shape sorter, stack blocks distractor, stack blocks target \\
\midrule
Control and Operation Tasks
  & cream cheese, butter, desk caddy, wine rack, bulb, bulb holder, charger, lamp base, lamp screw, lampshade, light bulb, oven knob, waterer \\
\midrule
Cooking and Kitchen Operations
  & flat stove, chefmate frypan, moka pot, wooden tray, chopping board, oven tray, place item, task wall \\
\bottomrule
\end{tabular}
\end{table}

\subsubsection{Simulator Information}

As shown in Table \ref{tab:RLBench Objects}, we systematically evaluate five categories of scenarios on the RLBench and LIBERO platforms, covering object manipulation and placement, container and switch operations, stacking and assembly, control and operation, as well as typical kitchen cooking scenarios. In RLBench, these five categories cover 70 tasks involving 107 object classes. In LIBERO, they cover 130 tasks involving 37 object classes.

\begin{algorithm}[h]
\caption{Scene Graph Generation}
\label{alg:gripper_state}
\DontPrintSemicolon
\SetAlgoVlined

\textbf{Input:} Joint values $J$, gripper values $G$, and 3D pose
$P[i]=\bigl(x_{\min},x_{\max},y_{\min},y_{\max},z_{\min},z_{\max}\bigr)$,
where $i\in\{A,B,C,\dots\}$ indexes objects.\;

\textbf{Output:} Scene Graph:\;
\Indp
\parbox[t]{\linewidth}{\raggedright
``nodes'':[objects],\\
``edges'':[Robot State, Spatial Relations, Object States]
}\;
\Indm
\vspace{-10pt}  
\textbf{Definitions:}\;
\Indp
$J_{\tau}$: Threshold for joint opening state.\;
$G_{\tau}$: Threshold for gripper opening state.\;
$Vol(A)$: Volume of object A.\;
$Vol(B)$: Volume of object B.\;
$Vol(A \cap B)$: Intersection volume between A and B.\;
\Indm

\ForEach{$i \in \mathcal{O}$ \textbf{with} $(J,G,P_i)$}{
    \If{$G > G_{\tau}$}{
        Gripper state $\leftarrow$ Grasping\;
    }

    \If{$J > J_{\tau}$}{
        Object state $\leftarrow$ Open\;
    }\Else{
        Object state $\leftarrow$ Closed\;
    }

    \If{$\text{Intersection of A and B Volume, IoA} > \tau$}{
        A Inside B; B Contains A\;
    }

    Calculate Z-axis: $Z_A = \text{pose\_A}[2], \, Z_B = \text{pose\_B}[2]$\;

    \If{$Z_A < Z_B$}{
        A Under B\;
    }\Else{
        A OnTop B\;
    }
}
\textbf{return SG} 
\end{algorithm}

\subsubsection{Scene Graph Generation in Simulation} \label{subsubsection:SSG}
\begin{figure*}[h!]
  \centering
  \includegraphics[width=1.0\linewidth]{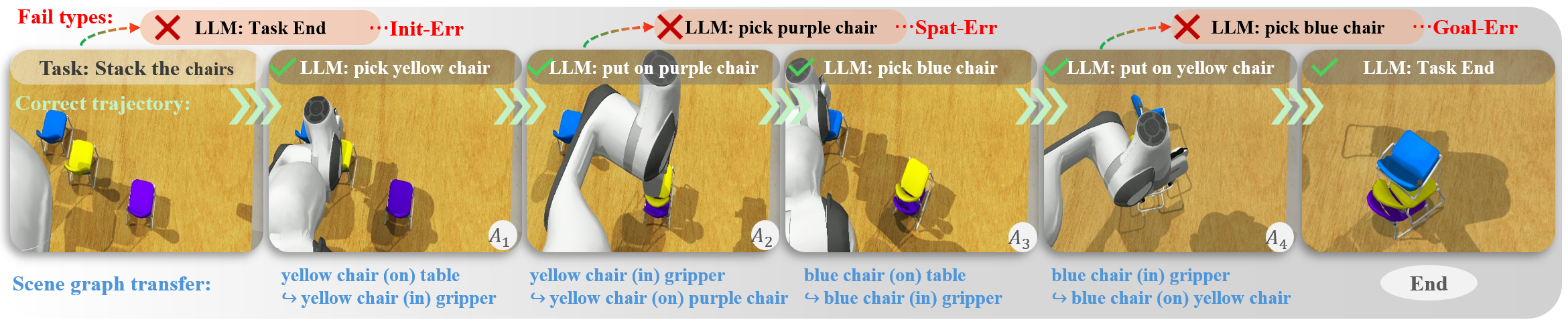}
  \caption{Task Cases in SA. stack the chairs. Correct task completion process and model-generated erroneous instructions.}
  \label{fig:stack_chairs-1}
\end{figure*}
To enable LLMs to comprehend scene information, we introduce a scene graph to represent spatial relations after robotic manipulation. As shown in Table \ref{tab:obj state}, we use gripping states to express robotic claw positions and employ Open and Close, Turn on and Turn off to denote intrinsic object states.
In the scene graph computation module, we determine spatial relations between independent objects—such as "Under" and "OnTop" —based on their intersection over union (IOU) and a predefined "relation reshold". For objects with jointed connections (e.g., cabinet doors, drawers), we compute joint angles or relative displacements between sub-objects and main objects. Examples include the angle between a cabinet door and the cabinet body (derived from simulator joint angle interfaces or calculated via BBOX), and the relative displacement between a single drawer and the cabinet body (derived from simulator joint values or calculated via BBOX). Object relations and states can be computed using via BBOX IOU. Combined with 6D pose detection in open-world environments, this enables rapid deployment in real-world settings and real-time scene graph generation. For the specific algorithm flow, see Algorithm \ref{alg:gripper_state}

\begin{algorithm}[htbp]          
\caption{Action Execution Engine}
\label{alg:compact_action_engine}
\DontPrintSemicolon
\SetAlgoVlined

\textbf{Input:} Action $a\in\mathcal{A}$, target object $o\in\mathcal{O}$\;
\textbf{Output:} Success flag $\sigma$, final scene graph $\mathcal{SG}^*$, verification log $\mathcal{V}$\;

$\mathcal{SG}\gets\mathrm{InitSceneGraph}()$, \quad $\mathcal{V}\gets\emptyset$\;
\If{$o\notin\mathcal{O}(\mathcal{SG})$}{\Return $(0,\mathcal{SG},\mathcal{V})$}
$\mathcal{W}\gets\mathrm{RetrieveWaypoints}(o,\mathcal{SG})$\;
$W^*\gets\{W\in\mathrm{FilterByAction}(\mathcal{W},a)\mid \mathrm{Feasible}(W,\mathcal{SG})\}$\;
\If{$W^*=\emptyset$}{\Return $(0,\mathcal{SG},\mathcal{V})$}
$W\gets\mathrm{Select}(W^*)$\;

\ForEach{$w_i\in W$}{
    $d_i\gets\|\mathbf{x}_r-\mathbf{x}_{w_i}\|_2$\;
    $\mathcal{P}_i\gets
    \begin{cases}
        \mathrm{InterpPlan}(\mathbf{x}_r,\mathbf{x}_{w_i}), & d_i\le\delta\\
        \mathrm{RRTPlan}(\mathbf{x}_r,\mathbf{x}_{w_i}),   & \text{otherwise}
    \end{cases}$\;
    \If{$\mathcal{P}_i=\emptyset$}{\Return $(0,\mathcal{SG},\mathcal{V})$}
    $\mathcal{SG}\gets\Phi(\mathcal{SG},\mathrm{Execute}(\mathcal{P}_i))$\;
    $\mathcal{SG}\gets\Phi(\mathcal{SG},\mathrm{ApplyGripper}(w_i,a))$\;
    $v_i\gets\psi_a(o,\mathcal{SG})$, \quad $\mathcal{V}\gets\mathcal{V}\cup\{(i,v_i)\}$\;
    \If{$v_i=0$}{\Return $(0,\mathcal{SG},\mathcal{V})$}
}
$\mathcal{SG}^*\gets\mathrm{FinalizeSceneGraph}(\mathcal{SG})$\;
\Return $(1,\mathcal{SG}^*,\mathcal{V})$
\end{algorithm}

\subsubsection{High-level Control interface}
 \label{subsubsection:HCI}
To enable motion planning based on LLM instructions within the RLBench and LIBERO environments, we build a robot control interface. 
For each object in the environment, we define multiple operation paths. 
Using the object coordinate system as a reference, we design distinct interaction methods (by setting a series of different path points). 
For instance, for a cup, we provide operations like full-body grasping, handle grasping, and pouring. 
Each operation has a corresponding approach method designed within the object coordinate system. 
Simultaneously, the system acquires the target object's real-time 6D pose and transforms the target point and approach trajectory from the object coordinate system to the robot coordinate system.
Finally, the system employs the classical RRT* algorithm to solve the complete path from the robot's current position to the target trajectory. It then calculates joint values using IK to generate path planning and control commands. The underlying controller executes the robot control tasks, ensuring precise motion execution.

\subsubsection{Command Interface}
To integrate the motion planning module with LLMs, we develop an upper-level action command system enabling direct interaction with the underlying motion control. This system uses a standardized command format: "action type + target object" (e.g., pick apple, place on table). The system then determines the robot arm's required posture for manipulating the target object by matching the action type in the command, and specifies the gripper control method.
For example, as shown in Figure \ref{fig:stack_chairs-1}, the current best model can execute and complete the "stack the chairs" task  in SA with the correct sequence: “LLM: pick yellow chair, LLM: put on purple chair, LLM: pick blue chair, LLM: put on yellow chair”. However, Claude-4-5 performs poorly on SA-type tasks, generating incorrect action sequences such as “LLM:pick purple chair” or “LLM:pick yellow chair, LLM:put on blue chair”.

\clearpage
\subsection{Details of GSR-Bench} \label{Appendix:GSR}
\subsubsection{Simulation Infrastructure}
GSR-Bench is built upon a lightweight, web-native simulation framework designed specifically for evaluating high-level embodied reasoning, spatial understanding, and decision-making. Unlike traditional simulators that focus on low-level motor control,  GSR-Bench emphasises the validation of an agent's ability to interpret complex scenes and plan long-horizon tasks.

To support this, the benchmark provides a rich set of 63 distinct object types, ranging from simple geometric primitives to complex articulated furniture. These objects are interconnected through a set of spatial and logical relations (e.g., OnTop, Inside, IsOpened) that form the basis of the scene graph. A summary of the detailed asset library and defined state space can be found in Table \ref{tab:gsr_assets}.

\begin{table}[htbp]
\centering
\caption{GSR-Bench Asset Library}
\label{tab:gsr_assets}
\begin{tabularx}{\columnwidth}{@{}l c X@{}}
\toprule
\textbf{Category} & \textbf{Count} & \textbf{Typical Examples} \\ \midrule
\textbf{Basic Primitives} & 11 & Boxes (Red, Yellow, Blue), Mugs (Red, Yellow, Blue), Cubes (Red, Yellow, Blue, Green, White) \\ \addlinespace
\textbf{Articulated Assets} & 10 & Movable furniture and appliances, including Cabinets with multiple drawers and Refrigerators, and Boxes with functional lids. \\ \addlinespace
\textbf{Stable HOPE} & 14 & Standardized food and condiment items: Milk cartons, Ketchup, and Mustard bottles. \\ \addlinespace
\textbf{Stable Scanned} & 11 & High-fidelity scanned kitchenware: Plates, Bowls (White/Red), and Trays. \\ \addlinespace
\textbf{Turbosquid} & 17 & Diverse commercial 3D models: Coffee machines and specialized kitchen appliances. \\ \midrule
\textbf{Total Assets} & \textbf{63} & A diverse collection for evaluating high-level task planning and decision-making. \\ \bottomrule
\end{tabularx}
\end{table}

And the technical architecture of GSR-Bench is designed to bridge high-level semantic planning with physically environment states through the following three components:

\begin{itemize}
    \item \textbf{Physics and Rendering}: The simulator employs cannon-es for rigid-body dynamics (gravity, friction, and collisions) to ensure physical plausibility. Three.js provides WebGL rendering, offering a high-throughput, browser-based alternative to heavy simulators like Isaac Gym for large-scale data synthesis.
    \item \textbf{Communication and Action Interface}: Communication and Semantic Interface: ROS2 serves as the communication middleware via rosbridge suite. ROSManager orchestrates the system by broadcasting a 10Hz State Summary and listening for high-level semantic commands (e.g., move [A] to [B]). To focus on reasoning, the simulator directly updates object states upon receiving commands, effectively decoupling task planning from robot kinematics.
    \item \textbf{Real-time Scene Graph Extraction}: Scene graphs are dynamically generated from the ground-truth physics state. Spatial relations (e.g., OnTop, Inside) are computed via AABB (Axis-Aligned Bounding Box), while logical states (e.g., $is\_opened$) are extracted directly from articulated object attributes.
\end{itemize}

\begin{figure}
    \centering
    \includegraphics[width=1\linewidth]{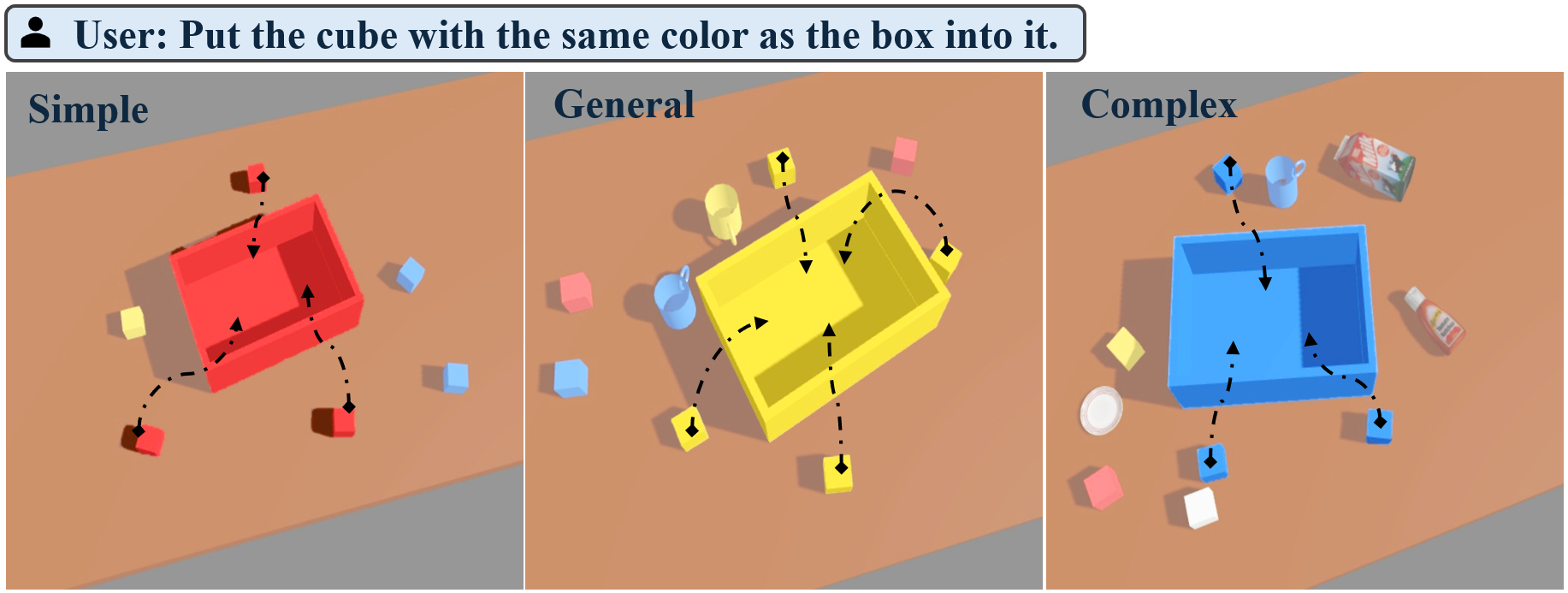}
    \caption{Examples of semantic object disambiguation tasks}
    \label{fig:GSR-task1}
\end{figure}

\begin{figure}
    \centering
    \includegraphics[width=1\linewidth]{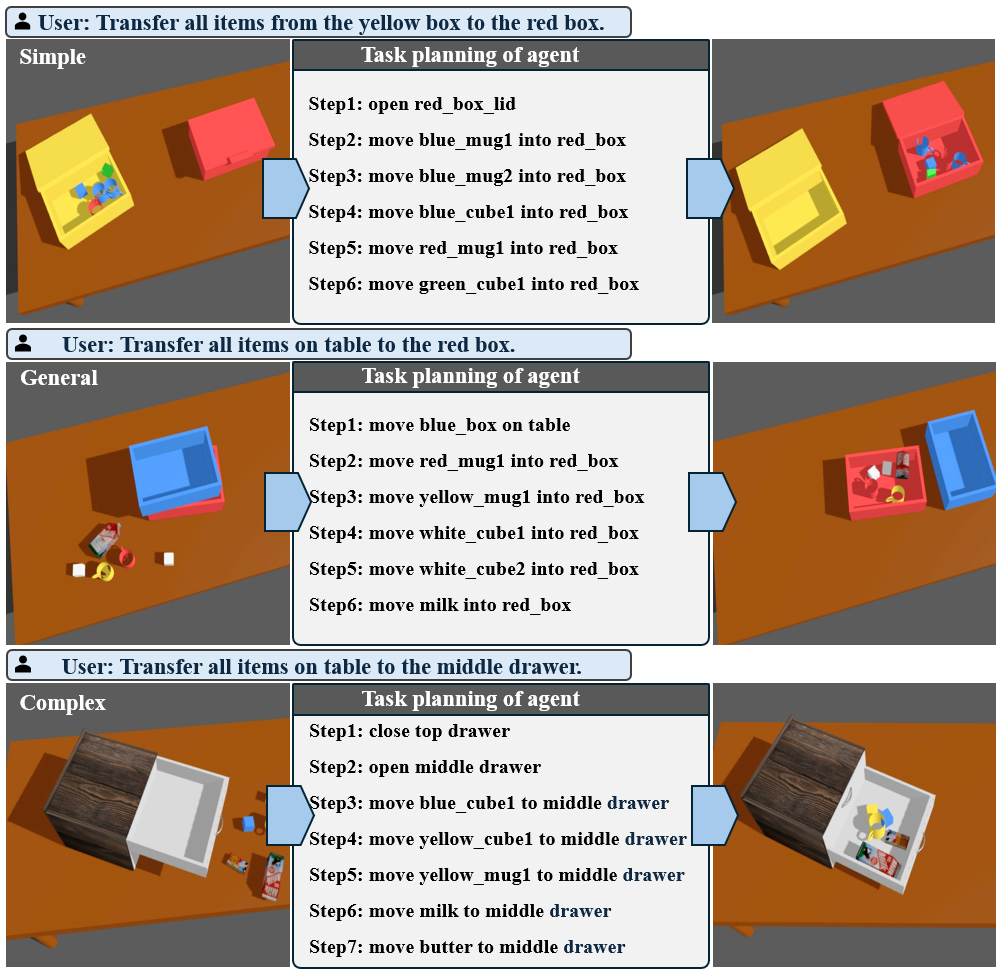}
    \caption{Examples of spatial-aware sequencing tasks}
    \label{fig:GSR-task2}
\end{figure}

\begin{figure}[htbp]
    \centering
    \includegraphics[width=1\linewidth]{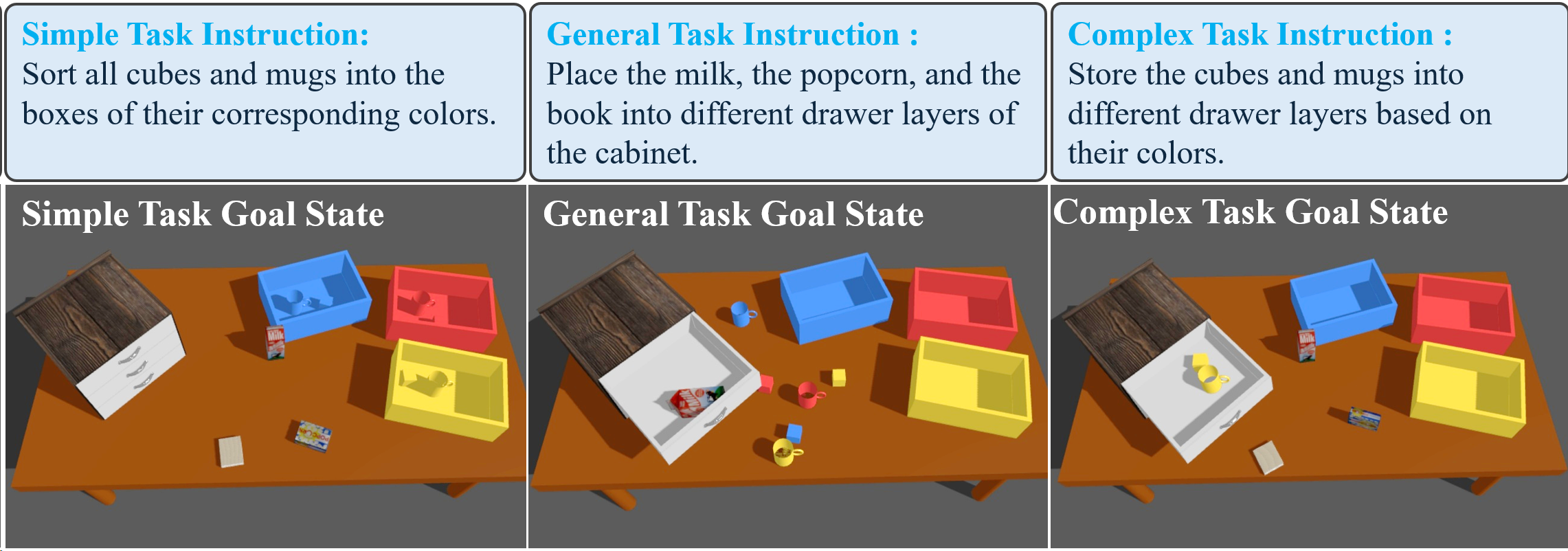}
    \caption{Examples of goal-conditioned generalization tasks}
    \label{fig:GSR-task3}
\end{figure}

\subsubsection{Task Suites}
GSR-Bench presents a comprehensive evaluation scheme intended to assess an embodied agent's performance across three fundamental cognitive domains: semantic object disambiguation, spatial-aware sequencing, and goal-conditioned generalization. Each suite comprises three levels of increasing difficulty, which are achieved by scaling up the number of obstacles, increasing physical constraints or broadening the scope of the instructions. The specific natural language instructions for each test are summarized in Table \ref{tab:task_instructions}. The detailed environmental configurations for each suite are described below:

\paragraph{Semantic Object Disambiguation} This suite assesses an agent's ability to identify target nodes within scene maps containing diverse object categories. The complexity of the scenes increases with higher densities of interfering elements and wider object category diversity. 
\begin{itemize} 
    \item \textbf{Easy}: The scene contains one box (randomly Red, Yellow, or Blue) and 7--10 cubes of various colors on the table. \textit{Instruction: "Pick up all cubes that have the same color as the box and put them inside the box."} 
    \item \textbf{General}: The distractor density increases with 9--12 cubes and mugs of various colors added to the scene. \textit{Instruction: "Pick up all cubes that have the same color as the box and put them inside the box."} 
    \item \textbf{Complex}: In addition to the items at General level, there are three randomly selected items of a different type (e.g., plate, bowl, milk) are introduced. \textit{Instruction: "Pick up all cubes that have the same color as the box and put them inside the box."}
\end{itemize}

\begin{table}[htbp]
\centering
\caption{Instruction Specification across GSR-Bench Task Suites}
\label{tab:task_instructions}
\begin{tabularx}{\columnwidth}{@{}>{\raggedright\arraybackslash}p{2.8cm} l >{\raggedright\arraybackslash}X@{}}
\toprule
\textbf{Task Suite} & \textbf{Level} & \textbf{Natural Language Instruction} \\ \midrule
\multirow{3}{2.8cm}{Semantic Object Disambiguation} & Simple & \multirow{3}{=}{"Pick up all cubes that have the same color as the box and put them inside the box."} \\
 & General & \\
 & Complex & \\ \midrule
\multirow{3}{2.8cm}{Spatial-Aware Sequencing} & Simple & "Transfer all items from the yellow box to the red box." \\ \addlinespace
 & General & "Move all objects on the table into the red box." \\ \addlinespace
 & Complex & "Transfer all objects from the box into the middle drawer." \\ \midrule
\multirow{3}{2.8cm}{Goal-conditioned Generalization} & Simple & "Sort all cubes and mugs into the boxes of their corresponding colors." \\ \addlinespace
 & General & "Place the milk, the popcorn, and the book into different drawer layers of the cabinet." \\ \addlinespace
 & Complex & "Store the cubes and mugs into different drawer layers based on their colors." \\ \bottomrule
\end{tabularx}
\end{table}

\paragraph{Spatial-Aware Sequencing} This test suite is designed to evaluate an agent's comprehension of physical causal chains and goal instructions. Its focus lies in identifying and resolving physical obstacles to generate a coherent sequence of actions.
\begin{itemize}
    \item \textbf{Simple}: Features two closed boxes (Red and Yellow). The Yellow box contains 5--8 random items. The agent must infer the explicit sequence of opening both lids to access and store items. \textit{Instruction: "Transfer all items from the yellow box to the red box."}
    \item \textbf{General}: Features Red, Yellow, and Blue boxes in a stacked configuration where the Red box is always at the bottom. 5--8 random objects are placed on the table. The agent must analyze $OnTop$ relations to clear the stack before accessing the target box. \textit{Instruction: "Move all objects on the table into the red box."}
    \item \textbf{Complex}: The scene contains 5--8 random items on table and a three-tier cabinet where the top drawer is open. The agent must recognize that the open top drawer physically obstructs access to the middle tier. \textit{Instruction: "Transfer all objects from the box into the middle drawer."}
\end{itemize}

\paragraph{Goal-conditioned Generalization} This suite tests the agent’s flexibility in translating abstract, long-horizon goals into state-aligned actions. The environmental layout is held constant with 3 colored boxes, 6--10 cubes/mugs, a milk carton, popcorn, and a book, while the instruction complexity evolves. 
\begin{itemize} 
    \item \textbf{Simple}: Focuses on mapping specific unique items to distinct functional locations. \textit{Instruction: "Place the milk, the popcorn, and the book into different drawer layers of the cabinet."} 
    \item \textbf{General}: Focuses on attribute-based multi-object grouping and sorting. \textit{Instruction: "Sort all cubes and mugs into the boxes of their corresponding colors."} 
    \item \textbf{Complex}: Requires cross-asset logical mapping, aligning semantic attributes with environmental tiers. \textit{Instruction: "Store the cubes and mugs into different drawer layers based on their colors."} 
\end{itemize}

\end{document}